\documentclass[10pt,journal,compsoc]{IEEEtran}
\ifCLASSOPTIONcompsoc
  \usepackage[nocompress]{cite}
\else
  \usepackage{cite}
\fi
\usepackage[pdftex]{graphicx}

\usepackage{multirow}
\usepackage{xcolor}
\usepackage{booktabs}
\usepackage{amssymb}
\usepackage{enumitem}
\usepackage{enumerate}

\usepackage{amsmath}
\ifCLASSOPTIONcompsoc
  \usepackage[caption=false,font=footnotesize,labelfont=sf,textfont=sf]{subfig}
\else
  \usepackage[caption=false,font=footnotesize]{subfig}
\fi
\usepackage{amssymb}

\begin{document}
\title{Learning Relation Prototype from Unlabeled Texts for Long-tail Relation Extraction}

\author{Yixin Cao, Jun Kuang, Ming Gao$^\ast$, Aoying Zhou, Yonggang Wen, Tat-Seng Chua% <-this % stops a space
\thanks{$^\ast$Corresponding author. This paper is an extended version of the ICDE'2020 conference paper~\cite{kuang2019improving}.}\protect\\
\IEEEcompsocitemizethanks{\IEEEcompsocthanksitem Yixin Cao and Yonggang Wen are with School of Computer Science and Engineering, Nanyang Technological University, Singapore. Email: caoyixin2011@gmail.com and ygwen@ntu.edu.sg.%\protect\\

\IEEEcompsocthanksitem Jun Kuang and Aoying Zhou are with School of Data Science and Engineering, East China Normal University, Shanghai, China. Email: ayzhou@dase.ecnu.edu.cn.

\IEEEcompsocthanksitem Ming Gao is with the School of Data Science and Engineering, and KLATASDS-MOE in the School of Statistics, East China Normal University, Shanghai, China, 200062. mgao@dase.ecnu.edu.cn.

\IEEEcompsocthanksitem Tat-Seng Chua is with School of Computing, National University of Singapore, Singapore. Email: dcscts@nus.edu.sg.

}% <-this % stops an unwanted space
}


\markboth{IEEE TRANSACTIONS ON KNOWLEDGE AND DATA ENGINEERING, SUBMISSION 2020}%
{Cao \MakeLowercase{\textit{et al.}}: Learning Relation Prototype for Long-tail Relation Extraction}



\IEEEtitleabstractindextext{%
\begin{abstract}
  Relation Extraction (RE) is a vital step to complete Knowledge Graph (KG) by extracting entity relations from texts. However, it usually suffers from the long-tail issue. The training data mainly concentrates on a few types of relations, leading to the lack of sufficient annotations for the remaining types of relations. In this paper, we propose a general approach to learn relation prototypes from unlabeled texts, to facilitate the long-tail relation extraction by transferring knowledge from the relation types with sufficient training data. We learn relation prototypes as an implicit factor between entities, which reflects the meanings of relations as well as their proximities for transfer learning. Specifically, we construct a co-occurrence graph from texts, and capture both first-order and second-order entity proximities for embedding learning. Based on this, we further optimize the distance from entity pairs to corresponding prototypes, which can be easily adapted to almost arbitrary RE frameworks. Thus, the learning of infrequent or even unseen relation types will benefit from semantically proximate relations through pairs of entities and large-scale textual information.
  
  We have conducted extensive experiments on two publicly available datasets: New York Times and Google Distant Supervision. Compared with eight state-of-the-art baselines, our proposed model achieves significant improvements (4.1\% F1 on average). Further results on long-tail relations demonstrate the effectiveness of the learned relation prototypes. We further conduct an ablation study to investigate the impacts of varying components, and apply it to four basic relation extraction models to verify the generalization ability. Finally, we analyze several example cases to give intuitive impressions as qualitative analysis. Our codes will be released later.

  %Relation Extraction (RE) is a paramount step to complete Knowledge Graph by extracting entity relations from texts. However, it usually suffers from the long-tail issue, as the training data mainly concentrates on a few types of relations, leading to the lack of sufficient annotations for the remaining types of relations. In this paper, we propose a general approach to learn relation prototypes from unlabeled texts, to facilitate the long-tail RE by transferring knowledge from those with sufficient data. We learn prototypes as an implicit factor between entities, to reflect the meanings of relations and their proximities. Specifically, we construct an entity co-occurrence graph from texts, and capture structural proximities for embedding learning. Furthermore, we optimize the distance from entity pairs to corresponding prototypes, which can be easily adapted to many RE framework. We have conducted extensive experiments on two publicly available datasets. Compared with eight state-of-the-art baselines, our model achieves significant improvements (4.1% F1 on average). Further results on long-tail relations demonstrate the effectiveness of the learned relation prototypes. We further conduct an ablation study to investigate the impacts of varying components and the generalization ability. Finally, we analyze several example cases to give intuitive impressions as qualitative analysis.

\end{abstract}

}

\maketitle

\IEEEdisplaynontitleabstractindextext

% \fi
\IEEEpeerreviewmaketitle

\IEEEraisesectionheading{\section{INTRODUCTION}\label{sec:intro}}

\IEEEPARstart{I}{n} the past decade, we have seen the emergence of various Knowledge Graphs (KGs), such as YAGO~\cite{yago} and DBPedia~\cite{dbpedia}. They have achieved great success in both academic and industrial applications, ranging from recommendation~\cite{cao2019unifying} to Question Answering~\cite{kbqa}. However, these KGs are far from complete, which limits the benefits of transferred knowledge. Relation Extraction (RE) is a vital step to complete KGs by extracting the relations between entities from texts. It is nontrivial since the same relation type may have various textual expressions, and meanwhile, different types of relations can also be described with the same words. Such ambiguity between relations and texts challenges the supervision of RE models.

\begin{figure*}[htb]
  \centering
  \subfloat[\label{fig:example_a}]{
    \includegraphics[width=0.45\linewidth]{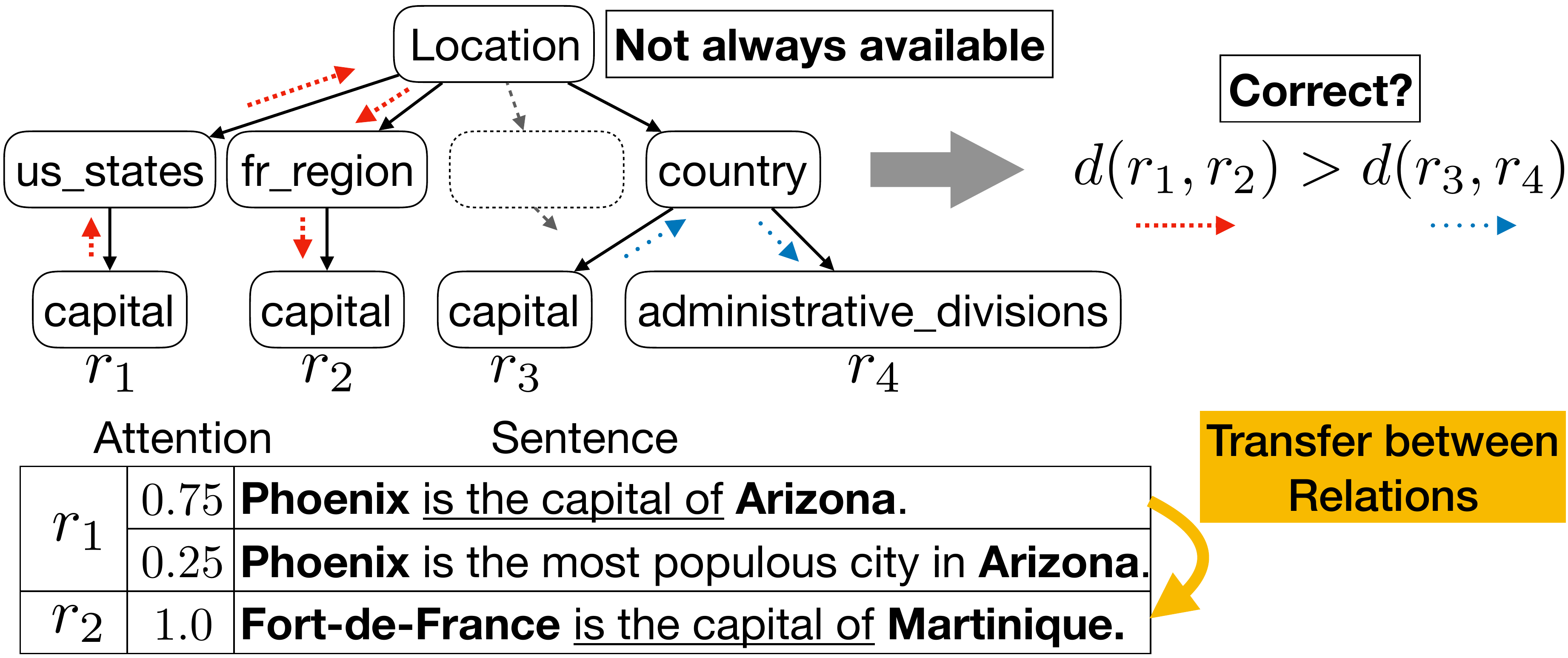}}
  \subfloat[\label{fig:example_b}]{
    \includegraphics[width=0.25\linewidth]{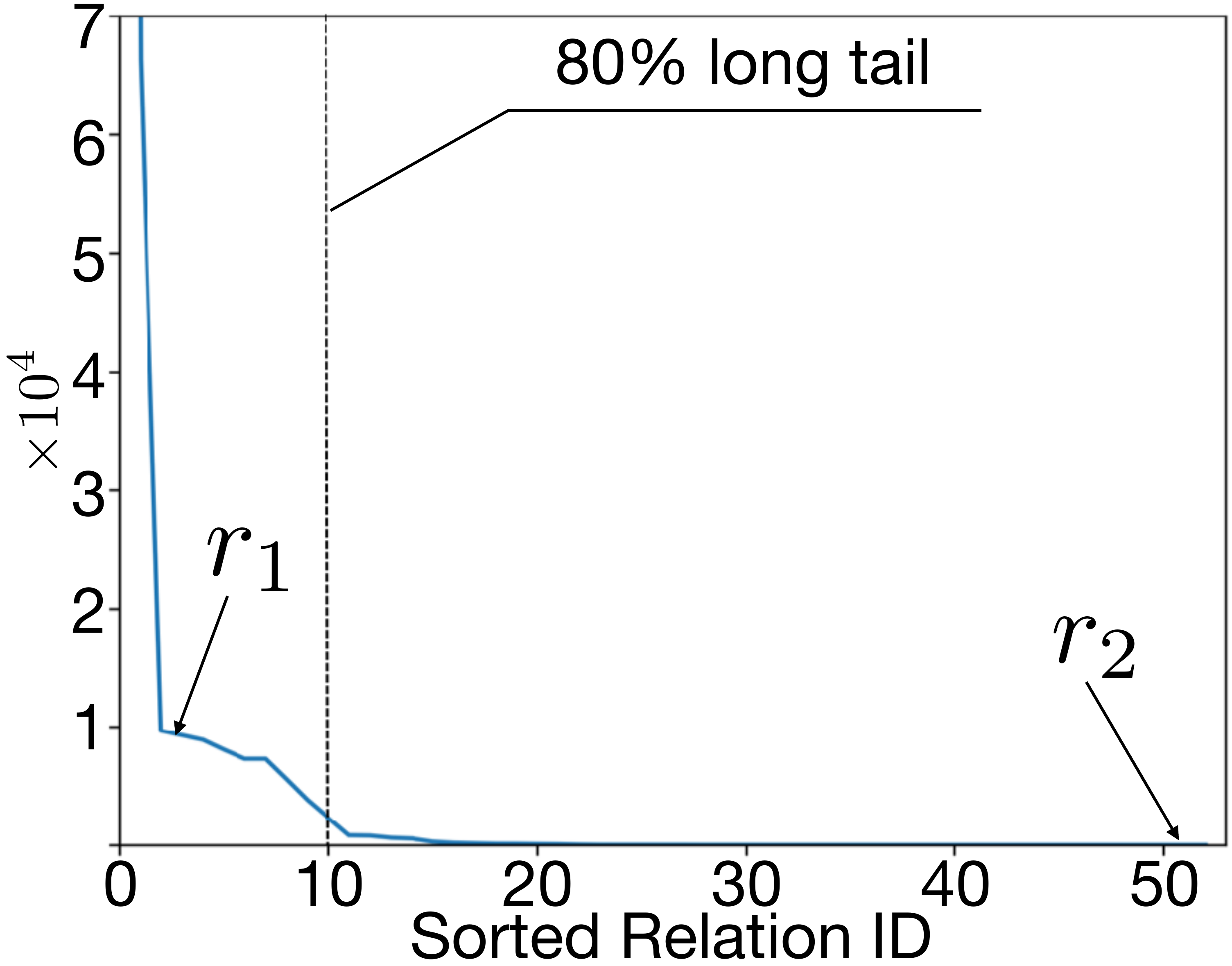}}
  \subfloat[\label{fig:example_c}]{
    \includegraphics[width=0.25\linewidth]{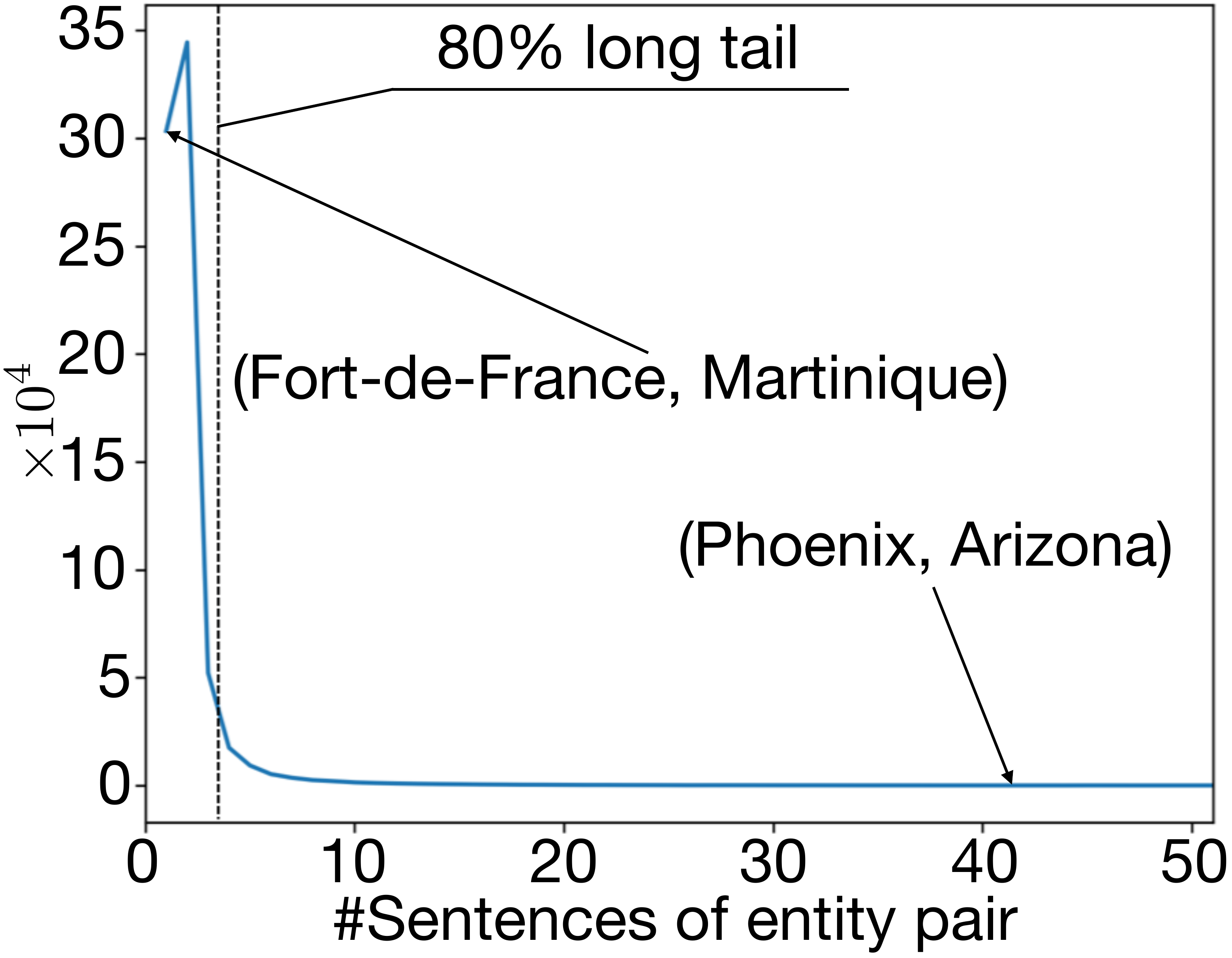}}
  \caption{Illustration of relations, training sentences, and their long-tail distributions in NYT. We present relation hierarchy as the widely used prior knowledge for long-tail RE, where dashed lines denote a distance measurement (Lowest Common Ancestor) between two relation types. We highlight similar textual patterns with underlines between proximate relations. We denote entities in sentences with bold fonts.}
	\label{fig:example}
\end{figure*}

Due to the expensive human annotation cost, distant supervision is proposed to automatically annotate the mappings between sentences and relations~\cite{distiawan2019neural}. It assumes that if two entities participate in a relation, a.k.a., a triple $(e_h,r_i,e_t)$ (i.e., \textit{head entity}, \textit{relation}, \textit{tail entity}) exists in KG, all of the sentences that contain $e_h$ and $e_t$ might express the relation $r_i$. However, it is argued that the quality and quantity of automatic annotations are usually not satisfactory~\cite{senatt,zhang2019long}. In terms of quality, much noise comes with the failure of the assumption --- some sentences include the same entity pair $(e_h,e_t)$ but express another relation $r_j$. As shown in Figure~\ref{fig:example_a}, given the triple (\textit{Phoenix}, \textit{/location/us\_state/capital}, \textit{Arizona}), we collect two sentences that include the entity pair (\textit{Phoenix}, \textit{Arizona}). Clearly, the first sentence expresses a similar meaning with the given relation type, but the second one implies another type of relation \textit{city of}, which brings in noise to the training corpora\footnote{As the term relation can refer to either relation type or relation instance (between specific entity pairs), in the paper, we simplify the use of term relation for relation type unless otherwise stated.}. To highlight informative sentences, many existing works introduce the attention mechanism to assign sentences with different learning weights~\cite{senatt}.

In terms of quantity, on the other hand, most of the training data collected by distant supervision concentrate mainly on a few relations, leading to the issue of the lack of sufficient annotations for the remaining relations. Take the widely used dataset, New York Times (NYT)~\cite{riedel}, as an example, we present the number of training instances of each relation in Figure~\ref{fig:example_b}. Unsurprisingly, the annotations are long-tail concerning different relations, and the tail relations suffer from insufficient training corpora. More specifically, each relation $r_i$ refers to multiple entity pairs $(e_h,e_t)$, and the sentences for each entity pair also have long-tail distributions (Figure~\ref{fig:example_c}). That is, only a small portion of entities frequently co-occur in sentences, then we can collect sufficient training data. Most entities rarely co-occur or are even unseen for the corresponding relations, which exacerbates the long-tail issue in relation extraction.

In this paper, we propose to improve relation extraction by learning prototypes for each relation type, which can be utilized to transfer knowledge from the relations with sufficient training data to the long-tail relations. As shown in Figure~\ref{fig:example_a}, the long-tail relation \textit{/location/fr\_region/capital} is semantically similar to the relation \textit{/location/us\_state/capital}, which has 798 training instances. They share a common textual pattern in sentences (e.g., \textit{is the capital of}), and thus we can utilize such relation proximity for transfer learning.

A major challenge lies in correctly identifying the proximity among relations; otherwise, the knowledge transfer between irrelevant relations will bring in noise and mislead the model training. Existing works~\cite{han2018hierarchical,zhang2019long} introduce the hierarchies of relations as prior knowledge. They assume that the smaller the structural distance between two nodes (e.g., Lowest Common Ancestor in Figure~\ref{fig:example_a}) is, the similar the relations are. However, such a prior hierarchy is not always available and sometimes incorrect. For instance, although the distance between $(r_3,r_4)$ is smaller than that between $(r_1,r_2)$, $(r_1,r_2)$ should be more similar with respect to RE prediction distributions because of the more common textual contexts. Therefore, how to capture relation proximity in a more precise and general way remains challenging.

Another major challenge is to distinguish between different relations, in case the knowledge transfer introduces a bias towards the same prediction for proximate relations. For example, as mentioned above, both \textit{/location/us\_state/capital} and \textit{/location/fr\_region/capital} indicate the capital relation, and the only difference is that between two United States entities or French entities. DPEN~\cite{gou2020dynamic} incorporates entity type information to learn relation-specific classifier dynamically. However, entity type information is sparse in KGs (nearly 40\% entities in DBPedia~\cite{dbpedia} do not have any type), challenging the scalability.

To address the first issue, we propose to learn relation prototypes that capture the proximity relationship among relations from involved entity pairs. Inspired by Prototypical Networks~\cite{snell2017prototypical}, we represent each relation prototype with the centroid of its training data, and each data point is defined as the difference between the pair of entity embeddings, namely implicit mutual relation (instance). Given any entity pair, we compute the implicit mutual relation and its distance to each relation prototype. These proximities suggest possible relations to the classifier, which further makes correct predictions by extracting discriminative signals from supportive sentences. Relation prototypes can also be enhanced by prior information (i.e., relation hierarchy and entity types), and be applied to arbitrary sentence encoder.

To address the second issue, we enhance entity embeddings with textual information for implicit mutual relation learning. In specific, we construct an entity co-occurrence graph from unlabeled texts and modeling both the first-order and second-order structural proximity. The massive textual contexts are helpful to infer entity types for distinguishment. Besides, long-tail entity pairs can also benefit from additional textual information.
We summarize our main contributions as follows:

\begin{itemize}[leftmargin=*]
    \item We highlight the importance of considering long-tail distributions of both relation types and their referred entity pairs (i.e., relation instances).
    
    \item We propose a novel model that learns relation prototypes from unlabeled texts, to improve long-tail relation extraction by transferring knowledge from relations with sufficient training instances. The learned prototypes can be applied to almost arbitrary RE models.
    
    \item We have conducted extensive experiments on two publicly available datasets, and compared them with eight state-of-the-art baseline methods. We have analyzed the impacts of relation prototypes, prior hierarchy, entity types, and implicit mutual relations, especially on long-tail settings. The results demonstrate the effectiveness of our proposed method.
\end{itemize}

A preliminary version of this work has been published in the conference of ICDE 2020~\cite{kuang2019improving}. We summarize the main changes as follows:

\begin{itemize}[leftmargin=*]
  \item Introduction (Section~\ref{sec:intro}). We have reconstructed the abstract and introduction to highlight the motivations of the extended version.
  \item Methods (Section~\ref{sec:solution}). We propose a novel relation prototype learning framework that extends the preliminary model to alleviate the long-tail issue in RE.
  \item Experiment (Section~\ref{sec:exp}). We add experiments to verify the new model on long-tail settings compared with another two baselines. We further explore the learned prototypes via ablation study and case study to justify the effectiveness and generalization ability.
  \item Preliminaries (Section~\ref{sec:pre}) and Related Work (Section~\ref{sec:rw}). The two sections are reconstructed to make the paper more complete and self-contained.
\end{itemize}

%The rest of the paper is organized as follows. In Section~\ref{sec:pre}, we formulate the problem and overview the framework, and Section~\ref{sec:solution} introduces our proposed method in detail. We report the promising experiment results on real-world datasets in Section~\ref{sec:exp}. Section~\ref{sec:rw} covers the related works. Finally, we conclude the paper in Section~\ref{sec:conclusion}.

\begin{figure*}[htp]
  \centering
  \includegraphics[width=0.9\linewidth]{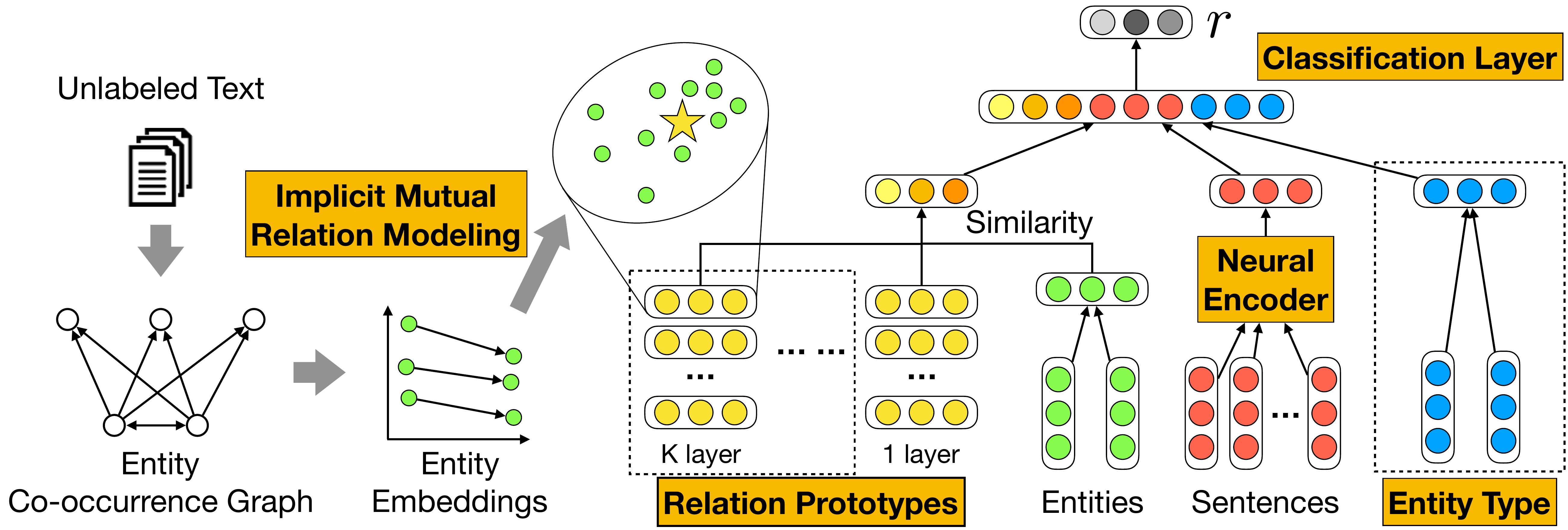}
  \caption{Framework. There are five main components: Relation Prototype Learning, Implicit Mutual Relations Modeling, Incorporation of Entity Type, Sentence Neural Encoder, and Classification Layer. We denote the same type of embeddings with the same color. Each relation prototype is the centroid (yellow star) of the implicit mutual relations (green circles) between referred pairs of entities. The components with dashed rectangles are optional. We omit the co-occurrence times in the graph for clarity.}
  \label{fig:framework}
  \end{figure*}

\section{PRELIMINARY}~\label{sec:pre}
In this section, we first formulate the task and necessary notations, and overview our framework.

\subsection{Notations}

\textbf{Knowledge Graph} is a directed graph $\mathcal{G}=(\mathcal{E},\mathcal{R},\mathcal{T})$ consisting of a set of entities $\mathcal{E}=\{e\}$, a set of relations $\mathcal{R}=\{r\}$ and a set of triples $\mathcal{T}=\{e_h,r_i,e_t\}$ that denote factual relationship between a pair of entities. Entities may have type information $c$ in the KG, such as \textit{person} and \textit{location}. Following conventions, we use bold-face letters to denote the embeddings of corresponding terms. For example, $\mathbf{e}$ is the embedding vector of entity $e$.

\textbf{Implicit Mutual Relation} $MR$ aims to reflect implicit relations between two entities. We thus define it as the distance between an entity pair in the vector space. This is inspired by analogy semantics of Word2vec~\cite{mikolov2013efficient}, the entity pairs with the same relation type should have similar implicit mutual relations (e.g., $MR_1=vec(Phoenix)-vec(Arizona)\approx MR_2=vec(Columbus)-vec(Ohio)$).

\textbf{Relation Prototype} $RP$ is learned for each relation type. They define a metric space, in which the distance between two prototypes refects their relation proximity. Different from implicit mutual relation, which is defined on each entity pair, each relation prototype refers to one relation type that may consist of multiple entity pairs.

\textbf{Relation Extraction} aims to classify the relation $r^*$ between two entities $(e_h, e_t)$ within a predefined set $R$, given a set of training sentences $S = \{s_1,s_2,\cdots,s_n\}$, where each sentence $s_i$ offers boundaries of entities. As shown in Figure~\ref{fig:example_a}, given the top two sentences and entity mentions (bold fonts), RE aims to predict the relation type $r_1$.

\subsection{Framework}

As illustrated in Figure~\ref{fig:framework}, our proposed method consists of five main components. Note that the prior relation hierarchy and entity types are optional (dashed rectangles).

\begin{itemize} [leftmargin=*]
  \item \textbf{Relation Prototype Learning}: Instead of heavily relying on prior hierarchy, we learn relation prototypes to reflect its meaning as well as the relationships with other relations. We achieve this based on implicit mutual relations. If the prior hierarchy is available, relations in different layers will have their own prototypes.
  
  \item \textbf{Implicit Mutual Relation Modeling}: Implicit mutual relations capture analogy semantics between entity pairs. We achieve this by (1) constructing an entity co-occurrence graph from unlabeled texts, and (2) modeling the structural proximity for entity embedding learning. MRs are also utilized to compute similarity with prototypes to estimate a preliminary relation for RE classifier.
  
  \item \textbf{Incorporation of Entity Type}: Entity types are commonly used to filter impossible relations between two entities. For example, \textit{Obama} is person entity, \textit{Hawaii} is a location, and their relation cannot be \textit{child of}. We learn type representations and feed them into the classifier, and other side information can also be incorporated in this way.

  \item \textbf{Sentence Neural Encoder}: RE models usually utilize a neural encoder to capture textual context information. This component is not our primary focus, so we adopt a widely used PCNN with attention to mitigate the negative impacts of noise. We also try different encoders to verify the generalization ability.
  
  \item \textbf{Classification Layer}: We introduce a classification layer to integrate the above four types of information and output a confidence score for each relation, indicating how possible the given entity pair and coupled sentences have the relation.

\end{itemize}

\section{METHODOLOGY}~\label{sec:solution}
For knowledge transfer, we learn prototypes for each relation type based on implicit mutual relation between each involved entity pair. We represent entity by pre-training their embeddings to capture both the first-order and second-order proximity over an entity co-occurrence graph, which is constructed using unlabeled texts only. Thus, the long-tail relations will benefit from their proximate relations with sufficient training corpora, and the infrequent or even unseen entity pairs will benefit from the large-scale textual information, leading to superior relation extraction performance.
In this section, we describe each component in detail.

\subsection{Relation Prototype Learning}
Relation Prototypes can capture relation type proximity and avoid the heavy reliance on prior hierarchy. Inspired by Prototypical Networks~\cite{snell2017prototypical}, we learn prototypes as centroids of training data, and data points are defined by implicit mutual relations between referred entity pairs:

\begin{equation}
  RP_i = \frac{1}{|\mathcal{T}_i|}\sum_{(e_h,r_i,e_t)\in \mathcal{T}_i}MR_{h,t}
  \label{equ:proto}
\end{equation}
where $RP_i$ is the prototype for relation $r_i$. We define $\mathcal{T}_i=\{(e_h,r_i,e_t)|e_h,e_t\in E,(e_h,r_i,e_t)\in \mathcal{T}\}$ as the triples with relation $r_i$, and $|\cdot|$ is the total number of the set. $MR_{h,t}$ is the implicit mutual relation between entities $e_h$ and $e_t$.

Typically, similar relations have similar implicit mutual relations, and the prototypes will be close in the vector space. Also, we can enhance the learned relation proximity with prior hierarchy if available. Assuming that the hierarchy has $K$ layers, we learn each node with a prototype:

\begin{equation}
  RP_i^k = \frac{1}{|\downarrow(r_i^k)|}\sum_{r_j^{k+1}\in \downarrow(r_i^k)}RP_j^{k+1}
  \label{equ:proto_hier}
\end{equation}
where $r_i^k$ denotes the $i$th node in the $k$th layer, and $\downarrow(r_i^k)$ denotes the set of children nodes. Note that all the prototypes for the leaf nodes $RP_i^K=RP_i$ can be computed according to Equation~\ref{equ:proto}. Thus, we can utilize the learned prototypes for transfer learning in Section~\ref{sec:class_layer}.

\subsection{Implicit Mutual Relations Modeling}

Implicit Mutual Relation aims to reflect the analogy semantics between entity pairs, such that $MR_1 = vec(Phoenix)-vec(Arizona) \approx MR_2 = vec(Columbus)-vec(Ohio)$, where these two pairs of entities have the same relation \textit{/location/us\_state/capital}. Meanwhile, the entity pair (\textit{Fort-de-France}, \textit{Martinique}) with a similar relation \textit{/location/fr\_region/capital} has a similar implicit mutual relation $MR'$. All of them reflect the \textit{capital} relationship and contribute common semantics to prototype learning.

Inspired by Word2vec~\cite{mikolov2013efficient}, we achieve this by capturing the co-occurrence among entities in texts. There are three stages for implicit mutual relations modeling. (1) We construct an entity co-occurrence graph based on the co-occurrence frequency of each entity pair. (2) Then, the entity representations are learned based on the entity co-occurrence graph. (3) We obtain the implicit mutual relations of any entity pair based on their representations.

\subsubsection{Entity co-occurrence graph construction}

\begin{figure}[htp]
\centering
\includegraphics[scale=0.4]{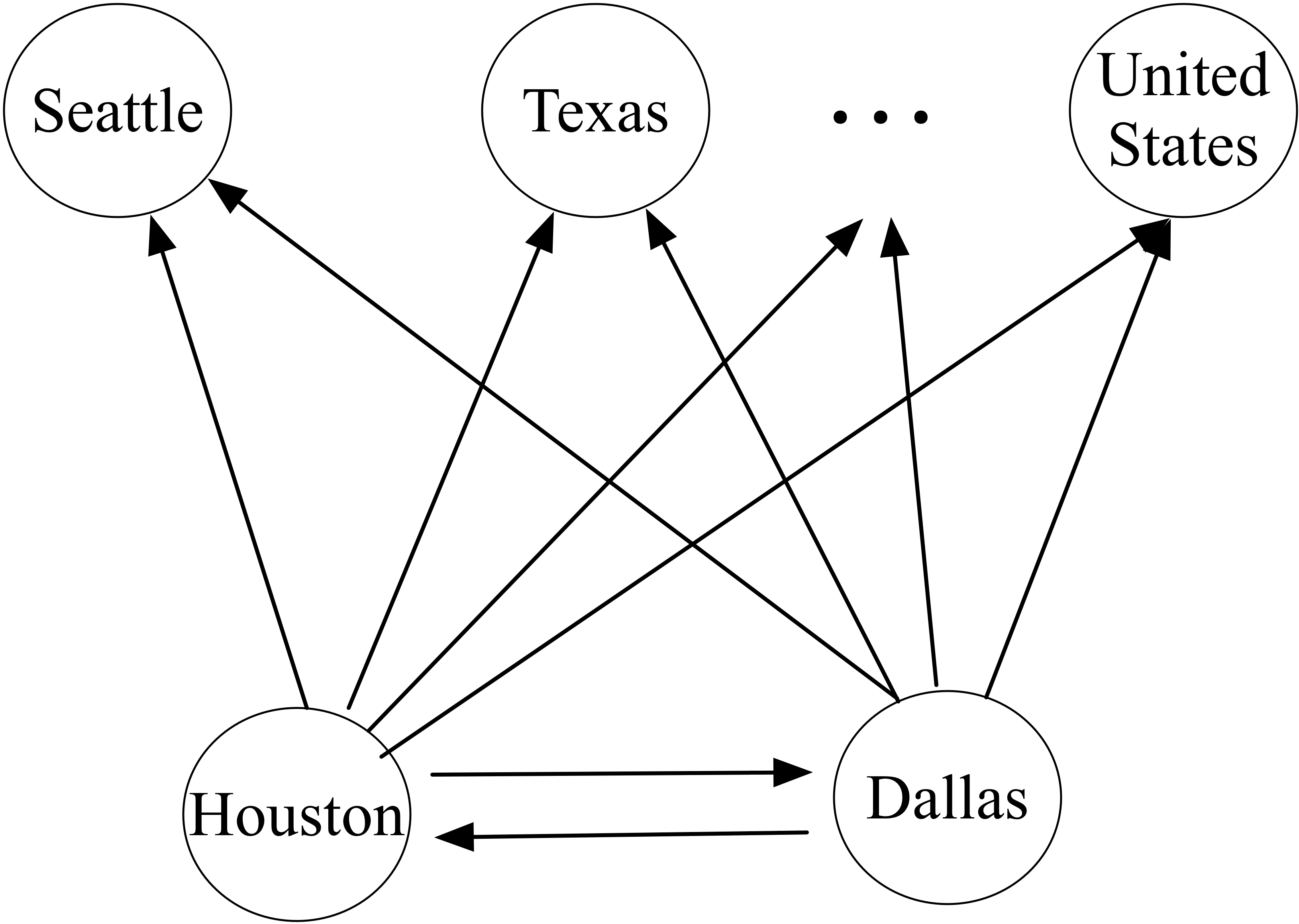}
\caption{The similar topological structure of entities \textit{Houston} and \textit{Dallas}, both which serve as locations to some relation (e.g., \textit{BornIn}).}
\label{fig:toplogical}
\end{figure}

The goal of entity co-occurrence graph is that entities in the graph play a similar role as the words in texts, which claim that words with similar contexts shall have similar embeddings. That is, entities with similar topological structures will have similar semantics. For example, as illustrated in Figure~\ref{fig:toplogical} (for clarity, we have omitted some unimportant points and edges), there are directed edges between entities \textit{Houston} and \textit{Dallas}. They are similar in semantics, and such similarity can be simply evaluated by the number of common neighbors between these two entities in the graph. 

To construct the graph, we utilize large scale unlabeled corpora, such as Wikipedia corpora, TIME magazine, and Google News. We employ exact string matching to identify entities from the unlabeled corpora, and then count the co-occurrences of entity pairs. For example, entities \textit{Obama} and \textit{Hawaii} exist in the same sentence ``Obama was born in Honolulu, Hawaii.'', then the co-occurrence time of \textit{Obama} and \textit{Hawaii} will increase 1. 

Each entity is a vertex in the entity co-occurrence graph, and each edge is assigned with a weight as follows:

\begin{equation}
    w_{i,j} = \frac{\log{(co_{i,j})}}{\log{(\max_{(e_k,r,e_l)\in \mathcal{T}}\{co_{k,l}\})}},
    \label{equ:weight}
\end{equation} 
where the value of $co_{i,j}$ denotes the co-occurrence number of entity pair $(e_i, e_j)$, and $\max_{(e_k,r,e_l)\in \mathcal{T}}\{co_{k,l}\}$ denotes max co-occurrence number among all entity pairs.
To reduce the computation complexity, we remove the entity pairs that co-occur below a manually defined threshold. In experiments, we have tried $\{10,20,30\}$ and found similar results. Thus, we set the thresholds to $10$ and $100$ for GDS and NYT datasets, leading to 15270(121218) and 69040(2085206) entities(relations), respectively. In the weighted graph, two vertices with similar topological structures indicate that the corresponding entities have similar semantics in unlabeled corpora. Next, we learn entity embeddings by preserving the structure information.

\subsubsection{Learning entity embedding}
The entity embeddings can be pretrained offline, which will not decrease RE efficiency.
Here, we follow a widely used network embedding approach~\cite{line} due to its effectiveness yet simplicity. More fancy models may improve embedding quality and RE performance. We do not explore them, because the goal of this paper is to provide a general framework for long-tail RE and the superior network embedding methods are out of the scope.
We consider the graph structure with first-order and second-order proximity. To be specific, \textit{first-order} proximity is defined to capture observed links in the graph, and the \textit{second-order} proximity aims to capture the contextual proximity between entities.

\textbf{First-order proximity:} A superior way to preserve the \textit{first-order} proximity is to minimize the distance between the joint probability of entity pair and its empirical probability. Following Line~\cite{line}, KL-divergence is chosen as distance measurement to minimize these two distributions. The final objective function is as follows:
\begin{equation}
    O_1 = - \sum_{e_i,e_j \in \mathcal{E}} w_{i,j}\cdot \log {P_1(e_i,e_j)},
\end{equation}
where the $w_{i,j}$ denotes the weight of edge between entities $e_i$ and $e_j$ defined in equation~(\ref{equ:weight}), and $P_1(e_i,e_j)$ denotes the joint probability between $e_i$ and $e_j$, which is defined as:

\begin{equation}
  P_1(e_i, e_j) = \frac{1}{1+exp(-\textbf{e}_i^T \cdot \textbf{e}_j)},
  \label{equ:jointprob}
\end{equation}
where $\textbf{e}_i \in \mathbb{R}^{d1}$ is the vector representation of entity $e_i$ in the $d1$-dimensional embedding space.

\textbf{Second-order proximity:} We assume that vertices with many shared neighbors are similar to each other, called the \textit{second-order proximity}. To preserve it, for each directed edge between $e_i$ and $e_j$ in the graph, we minimize the distance between the probability of context $e_j$ generated by vertex $e_i$ and its empirical probability. Similarly, when the KL-divergence is chosen, the objective function is as follows:

\begin{equation}
    O_2 = - \sum_{e_i,e_j \in \mathcal{E}} w_{i,j}\cdot \log{P_2(e_j|e_i)},
\end{equation}
where $P_2(e_j|e_i)$ is the probability of ``context'' $e_j$ generated by vertex $e_i$, which is defined as:

\begin{equation}
    P_2(e_j|e_i) = \frac{\exp{(\textbf{e}_j^T\cdot \textbf{e}_i )}}{\sum_{k=1}^{|\mathcal{E}|}{\exp{(\textbf{e}_k^T\cdot \textbf{e}_i )}}},
\end{equation}
where $|\mathcal{E}|$ denotes the total amount of vertices.

In practice, computation of the conditional probability $P_2(e_j|e_i)$ is extremely expensive. A simple and effective way is to adopt the negative sampling approach mentioned in~\cite{line}. Thus, the above objective function can be simplified:

\begin{equation}
O_2 = \log{\sigma(\textbf{e}_j^T \cdot \textbf{e}_i)} + \sum_{i=1}^K E_{e_n \sim N(e_i)} [\log{\sigma(-\textbf{e}_n^T \cdot \textbf{e}_i)}],
\end{equation}
where $N(e)$ denotes the neighbors of entity $e$ in the graph, $\sigma(x) = 1/(1 + exp(-x)) $ is the sigmoid function, and $K$ is the number of negative edges. The first term models the observed links, and the second term models the negative links drawn from the noise distribution.

To embed the vertices in the proximity graph, we preserve both the \textit{first-order} proximity and \textit{second-order} proximity separately, then obtain the embedding vector for a vertex by concatenating corresponding embedding vectors learned from the two models.

\subsubsection{Implicit mutual relation}
Considering that $vec(Phoenix)-vec(Arizona) \approx vec(Columbus)-vec(Ohio)$, we represent the implicit mutual relation between entities $e_i$ and $e_j$ as follows:

\begin{equation}
MR_{i,j} = \textbf{e}_j - \textbf{e}_i,
\end{equation}

\subsection{Incorporation of Entity Type}
We aim to propose a general way to deal with long-tail relation extraction, which can easily incorporate other side information. Here, we take entity type as an example, which is optional in our proposed model.
In intuition, entity types are beneficial to predict the relation. For example, \textit{/people/person/place\_of\_birth} is the relation between \textit{Location} and \textit{Person}, rather than \textit{Person} and \textit{Person}. Existing works~\cite{P94-100, P1183-1194, reside2018} have also shown that entity type information plays a positive role in relation extraction.

Following existing works~\cite{gou2019improving}, entity types are extracted from KG and further mapped to FIGER~\cite{P94-100}, while our proposed method is flexible in arbitrary entity type information. FIGER defines 112 fine-grained entity types. To avoid over-parameterization, we employ 38 coarse types in the first hierarchy in FIGER. Each entity type is embedded into $d2$ dimensional space to get the embedding vector of an entity type. When an entity has multiple types, we take the average over the embedding vectors.

We concatenate the type embeddings for a target entity pair $(e_i, e_j)$ as follows:

\begin{equation}
    \mathbf{c}_{i,j} = [\mathbf{c}_i || \mathbf{c}_j],
\end{equation}
where $\mathbf{c}_i$ is the type embedding for entity $e_i$, and $[\cdot || \cdot]$ denotes the concatenation operation.

\subsection{Sentence Neural Encoder}
To encode the supportive sentences, we introduce the widely used Piecewise CNN (PCNN) with sentence-level attention~\cite{senatt} as our neural encoder, although our proposed model can be applied with almost arbitrary encoders. The encoder can represent multiple sentences with a single fixed-length embedding, and highlight the informative sentence with higher attention weights. This component consists of three indispensable steps:

\begin{itemize} [leftmargin=*]
  \item[(1)] \textbf{Sentence Embedding}: Each sentence $s_i$ in a training sentence bag $S = \{s_1,s_2,\cdots,s_n\}$ should be represented by word embedding and relative position embedding, which means the relative position of all words in the sentence to the target entities. 

  \item[(2)] \textbf{Sentence Encoding}: As the previous works \cite{pcnn,senatt} shown, the convolutional neural networks with piecewise max-pooling (PCNN) is a fast and effective way to encode the sentence. Consequently, we get the encoding of each sentence using PCNN.
    
  \item[(3)] \textbf{Sentence-Level Attention}: The distant supervision suffers from noisy labels, i.e., not all sentences in a bag can express the target relation for the given entity pair. To address this issue, we utilize sentence-level attention to mitigate effects from noise sentences. To encode the bag of sentences, the model gives each sentence a score indicating the probability of expressing the relation. 
\end{itemize}

The encoding of the $i$th sentence bag can be represented:
\begin{equation}
    \mathbf{X}_{bag_i} = \sum_{j\in bag_i} \alpha_j \mathbf{x}_j,
\end{equation}
where $X_{bag_i}$ denotes the bag formed by all training sentences of the $i$th entity pair, and $\mathbf{x}_j$ is the embedding of sentence $s_j$. The score $\alpha_j$ for sentence $s_j$ is calculated by the selective sentence attention:
\begin{equation}
    \alpha _ { j } = \frac { \exp \left( q _ { j } \right) } { \sum _ { k } \exp \left( q _ { k } \right) },
\end{equation}
where $q_j$ is a query-based function which scores how well the sentence $s_j$ and the predict relation $r$ matches. We use the bi-linear function to calculate the scores:

\begin{equation}
    q _ { j } = \mathbf { x } _ { j } \mathbf { A } \mathbf { r },
\end{equation}
where $\mathbf{A}$ is a weighted diagonal matrix, and $\mathbf{r}$ is the query vector associated with relation $r$.

\subsection{Classification Layer}
\label{sec:class_layer}

Given a bag of sentences, relation prototypes, entity pairs $(e_h,e_t)$ and their types, the classification layer first takes each output of the above components as inputs, then outputs confident scores over predefined relation set $\{r_1,r_2,\cdots,r_m\}$. For inference, we choose the relation with the highest probability as the final prediction. For training, we choose cross-entropy loss:

\begin{equation}
  \mathcal{L} = - \sum \hat{y} log P(r|RP,e_h,e_t,S,c_h,c_t)
\end{equation}
where $\hat{y}$ is the groundtruth, and the probability distribution over $m$ relations between entities $e_h$ and $e_t$ can be computed:
\begin{equation}
    P(r|RP,e_h,e_t,S,c_h,c_t) = \sigma(W(\alpha C_{PR} + \beta C_{Type} + \gamma C_{RE})+b)
\end{equation}
where $\alpha$, $\beta$ and $\gamma$ are learnable weights of three following components $C_{RP}$, $C_{Type}$, and $C_{RE}$. In experiments, the ratio of them is around $1.5:1:0.6$ on average, which indicates the importance of our proposed relation prototypes as well as the effectiveness of the type information. In specific, first, we compute the distance from the implicit mutual relation of $(e_h,e_t)$ to each relation prototype, to suggest possible relations that reflect the proximity for transfer learning.

\begin{equation}
  C_{RP^k} = \sigma( [d(MR_{h,t}, RP_{1}^k) || \cdots || d(MR_{h,t}, RP_{mk}^k)) )
\end{equation}
where $\sigma$ is the softmax function, $mk$ is the number of relations in the $k$th layer if the prior relation hierarchy is avaiable, otherwise $k=1$ indicates there is only the predefined relation set, and $r_{mk}=r_m$. $d(\cdot,\cdot)$ is the distance measurement, and we use L2 distance in experiments. We concatenate $C_{RP^k}$ for different layers as the final prototype-based features $C_{RP}$. Thus, proximate relations will have similar features. During training, this will regularize the prediction distribution of long-tail relations similar to their proximate relations with sufficient training data, thereby transferring knowledge for better performance.

For entity types, we concatenate the type embeddings of a target entity pair and then use a fully connected layer with a Softmax to calculate the confidence score:

\begin{equation}
    C_{Type} = \sigma(W^c \mathbf{c}_{h,t} +b^c ),
\end{equation}
where the $W^c$ and $b^c$ are trainable parameters.

For the sentence bag, we calculate the confidence score:

\begin{equation}
    C_{RE} = \sigma(W^{RE} \mathbf{X}_{bag} + b^{RE}),
\end{equation}
where $W^{RE}$ and $b^{RE}$ are trainable parameters. This component can be replaced by any RE model.

\subsection{Discussion}
\label{subsection:discussion}

The relation prototypes can flexibly combine with various RE models as well as side information. We integrate them with typical PCNN with attention and the type information as described above. In experiments, we will verify the effectiveness of the prototypes using other CNN-based and RNN-based RE models, which present positive effects.

Instead of the given entity graph in existing KGs, such as Wikipedia hyperlinks and YAGO~\cite{yago}, we capture the entity co-occurrence from texts, because (1) under the opened-world assumption, existing KGs are far from complete and may not cover all relevant information about entities; and the missing links and multi-relational data may bring noise; (2) we are inspired by Word2vec~\cite{mikolov2013efficient}, which presents the expect analogy nature by modeling co-occurrence; and (3) although TransE~\cite{transE} can also model the relation between entities as a translation operation in the vector space, similar to Word2vec, the explicitly modeled relation $r$ may have a different distribution with that in texts, which is the target source of RE. Nevertheless, we are interested in incorporating KG as side information in the future.

\section{EXPERIMENTS}~\label{sec:exp}
In this section, we conduct comprehensive experiments to evaluate the performance of our proposed approach by comparing it with eight SOTA systems and six variants of our approach on two public datasets. Through empirical study, we aim at addressing the following research questions:

\begin{itemize}
    \item RQ1: How does our proposed approach perform compared with state-of-the-art RE approaches?
    \item RQ2: How do the relation prototypes and implicit mutual relations affect the RE model?
    \item RQ3: Could the relation prototypes and implicit mutual relations improve the performance of existing RE methods, such as GRU, PCNN, and PCNN + ATT? 
\end{itemize}

Besides, we conduct a case study, which qualitatively demonstrates the effectiveness of our proposed method.

\begin{table}[]
  \caption{The statistic on NYT and GDS datasets.}
  \centering
  \scalebox{1.0}{
  \begin{tabular}{|c|c|c|c|c|}
  \hline
  \textbf{Datasets} & \multicolumn{2}{c|}{\begin{tabular}[c]{@{}c@{}}\textbf{NYT}\\ (\#\textbf{Relations: 53})\end{tabular}} & \multicolumn{2}{c|}{\begin{tabular}[c]{@{}c@{}}\textbf{GDS}\\ (\#\textbf{Relations: 5})\end{tabular}} \\ \hline \hline
       & \#Sentences                             & \#EntityPair                             & \#Sentences                            & \#EntityPair                            \\ \hline
  Training    & 522,611                                 & 281,270                                    & 13,161                                 & 7,580                                     \\ \hline
  Testing     & 172,448                                 & 96,678                                     & 5,663                                  & 3,247                                     \\ \hline
  \end{tabular}
  }
  
  \label{table:dataset}
  \end{table}

\subsection{Experimental Settings}
\subsubsection{Datasets}

\begin{figure*}[htb]
    \centering
    \subfloat[PR curve on NYT dataset\label{fig:pr_NYT}]{
      \includegraphics[width=0.49\linewidth]{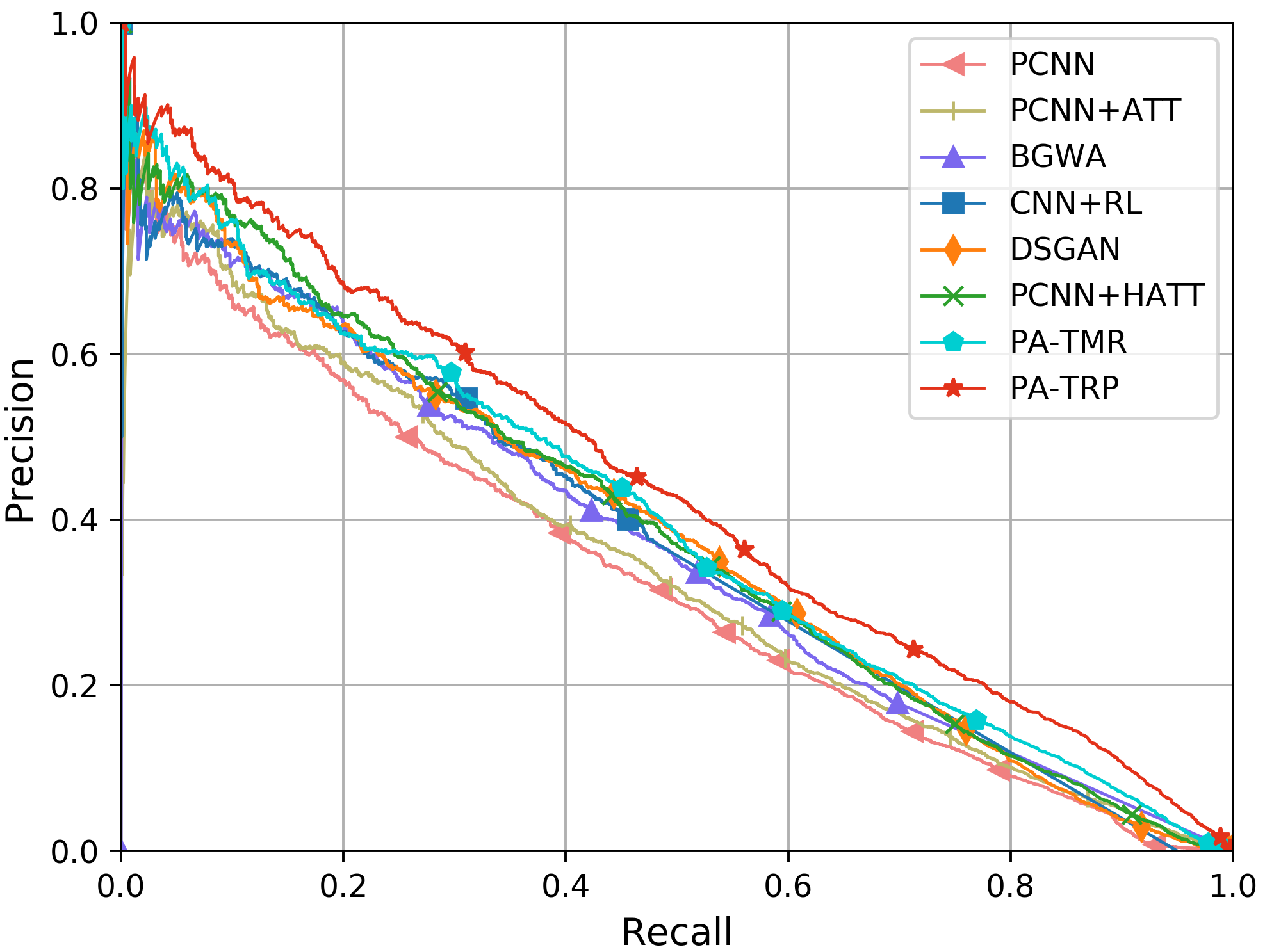}}
    \subfloat[PR curve on GDS dataset\label{fig:pr_GDS}]{
      \includegraphics[width=0.49\linewidth]{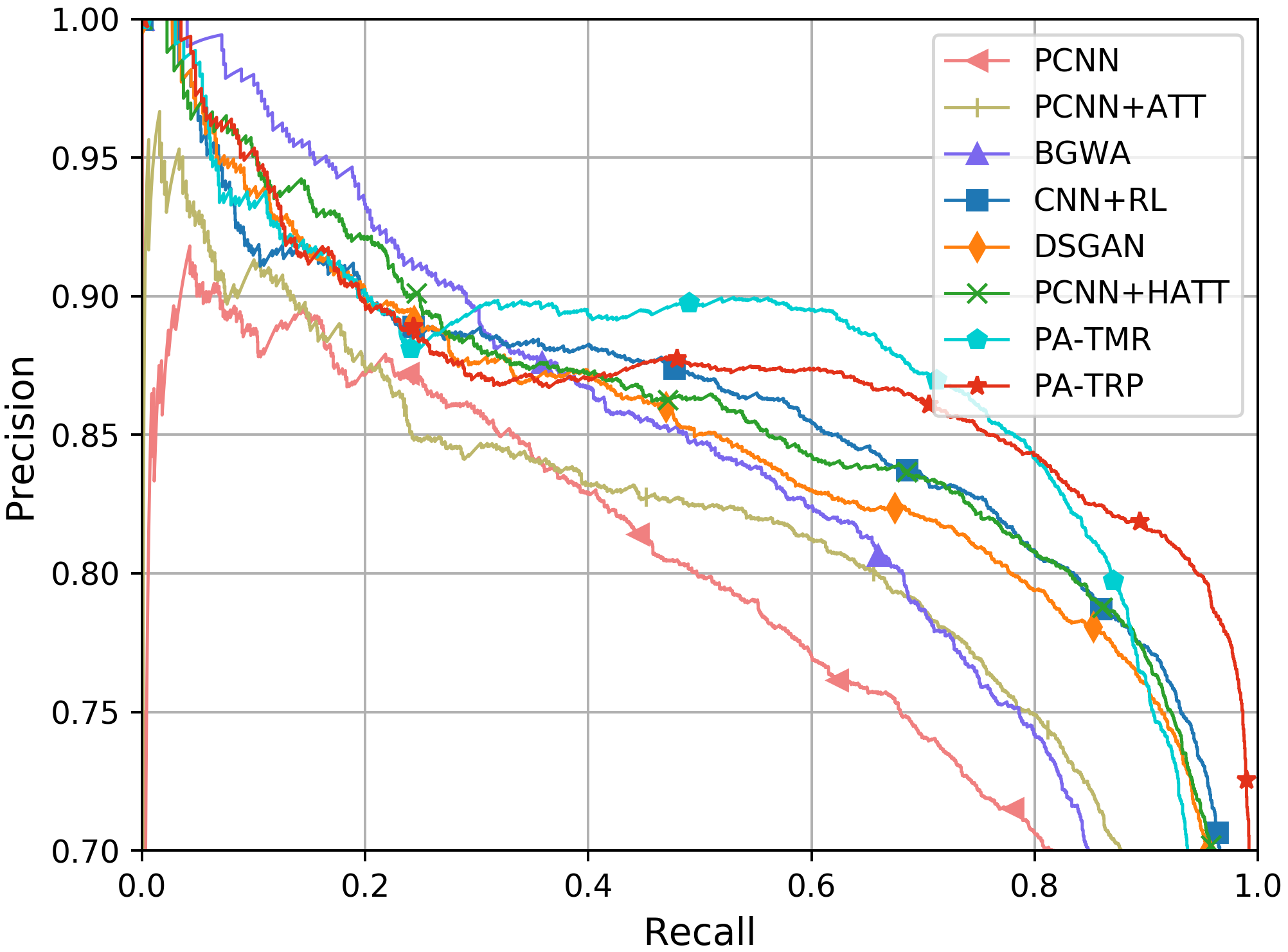}}
    \vspace{-0.1cm}
    \caption{The Precicion-Recall curves of six baseline methods as well as two variants of our methods on NYT and GDS datasets.}
    \vspace{-0.2cm}
      \label{fig:prcurve}
\end{figure*}

We adopt two widely used public datasets to demonstrate the effectiveness of our method and baselines. They are New York Times (NYT)~\cite{riedel} and Google Distant Supervision (GDS)~\cite{bgwa} datasets. The statistical descriptions of them are illustrated in Table~\ref{table:dataset}.

\begin{itemize} [leftmargin=*]
    \item NYT dataset is generated by annotating entities with Stanford NER tool in the New York Times corpus, which is then aligned with Freebase to get the relation between entities. The training samples are from the corpus of the years 2005-2006, and the testing samples are from the corpus of the year 2007. There are 53 different relations, including a relation NA that indicates there is no relation between two entities.

    \item GDS dataset is an extension of the manually annotated data set Google relation extraction corpus. The entities in Google relation extraction corpus are aligned with web documents to obtain new sentences, which also contain targeted entities. There are five different relations, including a relation NA.
\end{itemize}

To learn implicit mutual relations, we utilize Wikipedia articles as textual corpora to construct the entity proximity graph, since it thoroughly covers the entities existed in both NYT and GDS datasets. 

\subsubsection{Evaluation metrics}
Similar to most existing works, we evaluate our model with the held-out metrics, which compare the predicting relation facts from the test sentences with those in Freebase. We report the precision, recall, F1-score, precision at top $N$ prediction (P@N), and AUC (area under the Precision-Recall curve). For different thresholds, the precision and recall are different, so we report the precision and recall at the point of max F1-score.
In addition, we compute the average score for each metric after running the same experiment five times.

\subsubsection{Parameter settings}
We use grid search to tune the optimal model parameters. The grid search approach is used to select the learning rate $\lambda$ for stochastic gradient descent optimizer among $\{0.1,0.2,0.3,0.4,0.5\}$, the sliding window size $l$ of CNN among $\{1,2,3,4,5\}$, the number of filters $k$ of CNN among $\{180,200,230,250,300\}$, and the size of entity type embedding $d2$ among $\{10,15,20,25,30,40\}$. For the entity embedding size, we follow the setting of~\cite{line}. Table~\ref{table:para} shows the optimal parameters used in the experiments.

\begin{table}[]
  \caption{Parameter settings}
\centering
\begin{tabular}{|c|c|}
\hline
\textbf{Parameter Description}  & \textbf{Value}\\
\hline
Relation prototype embedding size & 128 \\
\hline
Entity type embedding size & 20  \\
\hline
Window size                     &3 \\
\hline
CNN filters number                 & 230 \\
\hline
POS embedding dimension           & 5   \\
\hline
Word embedding dimension          &50   \\
\hline
Learning rate                      & 0.3 \\
\hline
Sentence max length                & 120 \\
\hline
Dropout probability                 & 0.5 \\
\hline
Batch size                         & 160 \\
\hline
\end{tabular}
\label{table:para}
\end{table}

\begin{table*}[]
\caption{Overall performance on NYT and GDS datasets.}
\label{table:result}
\centering
\begin{tabular}{|c|c|c|c|c|c|c|c|}
\hline
\textbf{Dataset}     & \textbf{Method} & \textbf{AUC}    & \textbf{Precision} & \textbf{Recall} & \textbf{F1-Score} & \textbf{P@100} & \textbf{P@200} \\ \hline \hline
\multirow{9}{*}{NYT} & PCNN             & 0.3296          & 0.3830             & 0.4020          & 0.3923            & 0.77           & 0.72           \\ \cline{2-8} 
                     & PCNN+ATT         & 0.3424          & 0.3588             & 0.4564          & 0.4018            & 0.75           & 0.75           \\ \cline{2-8} 
                     & BGWA             & 0.3670          & 0.3994             & 0.4451          & 0.4210            & 0.76           & 0.74           \\ \cline{2-8} 
                     & CNN+RL           & 0.3735          & 0.4201             & 0.4389          & 0.4293            & 0.79           & 0.73           \\  \cline{2-8}
                     & DSGAN        & 0.3801 & 0.4251 & 0.4591 & 0.4414 & 0.80 & 0.78\\
                     
                     \cline{2-8}
                     & PCNN+HATT        & 0.3857 & 0.4313 & 0.4476 & 0.4393 & 0.80 & 0.79\\
                     
                     \cline{2-8} 
                     % & PA-T             & 0.3572          & 0.3779             & 0.4586          & 0.4143            & 0.78           & 0.72           \\ \cline{2-8} 
                     % & PA-MR            & 0.3635          & 0.4091             & 0.4410          & 0.4244            & 0.79           & 0.78           \\ \cline{2-8} 
                     & PA-TMR           & 0.3939 & \textbf{0.4320}    & 0.4615 & 0.4463   & 0.83  & 0.79  \\ \cline{2-8}
                     % & PA-RP            & 0.4225          & \textbf{0.4460}             &  0.4492          & 0.4476           & \textbf{0.93}           & \textbf{0.85}           \\ \cline{2-8}
                     & PA-TRP            & \textbf{0.4371}          & 0.4291             &  \textbf{0.4995}          & \textbf{0.4616}           & \textbf{0.88}           & \textbf{0.82}           \\ \hline \hline
\multirow{9}{*}{GDS} & PCNN             & 0.7798          & 0.6804             & 0.8673          & 0.7626            & 0.88           & 0.90           \\ \cline{2-8} 
                     & PCNN+ATT         & 0.8034          & 0.7250             & 0.8474          & 0.7814            & 0.94           & 0.93           \\ \cline{2-8} 
                     & BGWA             & 0.8148          & 0.7725             & 0.7162          & 0.8385            & 0.99           & \textbf{0.98}           \\ \cline{2-8} 
                     & CNN+RL           & 0.8554          & 0.7680             & 0.9132          & 0.8343            & \textbf{1.00}            & 0.96           \\ \cline{2-8}
                     & DSGAN        & 0.8445 & 0.7526 & 0.9115 & 0.8245 &0.99 & 0.97 \\
                     \cline{2-8} 
                     & PCNN-HATT             & 0.8540          & 0.7728             & 0.8979          & 0.8307            & 0.99           & 0.97           \\
                     
                     \cline{2-8}
                     % & PA-T             & 0.8512          & 0.7925             & 0.8969          & 0.8414            & 0.96           & 0.94           \\ \cline{2-8} 
                     % & PA-MR            & 0.8571          & 0.8011             & 0.8947          & 0.8453   & 0.97           & 0.94           \\ \cline{2-8}  
                     & PA-TMR           & 0.8646 &  \textbf{0.8058}    & 0.8641          & 0.8339            & \textbf{1.00}   & \textbf{0.98}  \\  \cline{2-8}
                     % & PA-RP            & 0.8640          & 0.7796             &  0.9160          & 0.8423          & 0.97            & 0.96           \\ \cline{2-8}
                     & PA-TRP            & \textbf{0.8725}          & 0.7964             &  \textbf{0.9553}         & \textbf{0.8686}           & \textbf{1.00}           & \textbf{0.98}           \\ \hline
\end{tabular}
\end{table*}

\subsubsection{Baselines}
We compare with the following baselines:

\textbf{BGWA}~\cite{bgwa} is a bidirectional GRU based relation extraction model. It focuses on reducing the noise from distant supervision by using hierarchical attentions.

\textbf{PCNN}~\cite{pcnn} is a CNN based relation extraction model which utilizes the piecewise max pooling to replace the single max pooling to capture the structural information between two entities.

\textbf{PCNN+ATT}~\cite{senatt} combines the selective attention over instances with PCNN, which is expected to dynamically reduce the weights of those noisy instances, thereby reducing the influence of wrong labeled instances.

\textbf{CNN+RL}~\cite{reinforce} contains two modules: an instance selector and a relation classifier. The instance selector chooses high-quality sentences with reinforcement learning. The relation classifier makes a prediction by the chosen sentences and provides rewards to the instance selector.

\textbf{DSGAN}~\cite{qin2018dsgan} utilizes an adversarial learning framework to learn a sentence level true-positive generator. The generator is used to filter the noise in the distant supervision dataset, in which way to obtain a cleaned dataset. Then the cleaned dataset is used to train a RE model. In their paper, the best results are produced by PCNN+ATT.

\textbf{PCNN+HATT}~\cite{han2018hierarchical} improves PCNN+ATT by utilizing prior relation hierarchy, which provides the proximity information for transfer learning. It computes selective attention within each layer in the hierarchy, and concatenates all of the layers for final classification.

\textbf{PCNN+KATT}~\cite{zhang2019long} improves PCNN+HATT by utilizing GCN to model relation hierarchy and introducing pretrained KG embeddings as external knowledge.

\textbf{DPEN}~\cite{gou2020dynamic} dynamically learns relation-specific classifier, which utilizes the entity types and relation classes to generate the classification parameters.

We carefully implement all the above baselines for our empirical study except for PCNN+KATT and DPEN, because we cannot find their released codes. We thus compare with them on the old-version NYT dataset in the long-tail settings, as they reported such performance in the original papers. For other methods, we have fairly tuned the hyper-parameters, and use the optimal hyper-parameter setting to report the average performance of five runs. Compared to the original papers, their performance of our implementation are not worse.

We also investigate several variants of our proposed method \textbf{PA-TRP} (\textbf{P}CNN-\textbf{A}TT + \textbf{T}ype + \textbf{R}elation \textbf{P}rototypes), to provide insights on the impacts of each main component. \textbf{PA-TMR} (\textbf{P}CNN-\textbf{A}TT + \textbf{T}ype + implicit \textbf{M}utual \textbf{R}elation) directly utilizes implicit Mutual Relation as entity pair representations, without the learning relation prototypes, which is proposed in the previous paper~\cite{kuang2019improving}. \textbf{PA-T} only adopts the type information, \textbf{PA-MR} only adopts implicit mutual relation, and \textbf{PA-RP} learns relation prototypes without the incorporation of entity types. In addition, to verify the effect of relation hierarchy, we remove the prior information in \textbf{PA-TRP/h}.

\subsection{Performance Comparison (\textbf{RQ1})}
\label{subsection:q1}

Figure~\ref{fig:prcurve} and Table~\ref{table:result} show the overall performance of our models as well as the baseline methods on both NYT and GDS datasets. First, from the precision-recall curves in Figure~\ref{fig:prcurve}, we can see that NYT is a more challenging dataset than GDS, and the testing performance is much more stable on NYT than that on GDS. This accords with the statistics in Table~\ref{table:dataset} that NYT has more relations than GDS to predict ($53$ v.s. $5$), and both the training and test sets are at large scale. We have the following key observations:

\begin{figure*}[htb]
  \centering
  \subfloat[NYT dataset\label{fig:f1_diff_nyt}]{
    \includegraphics[width=0.4\linewidth]{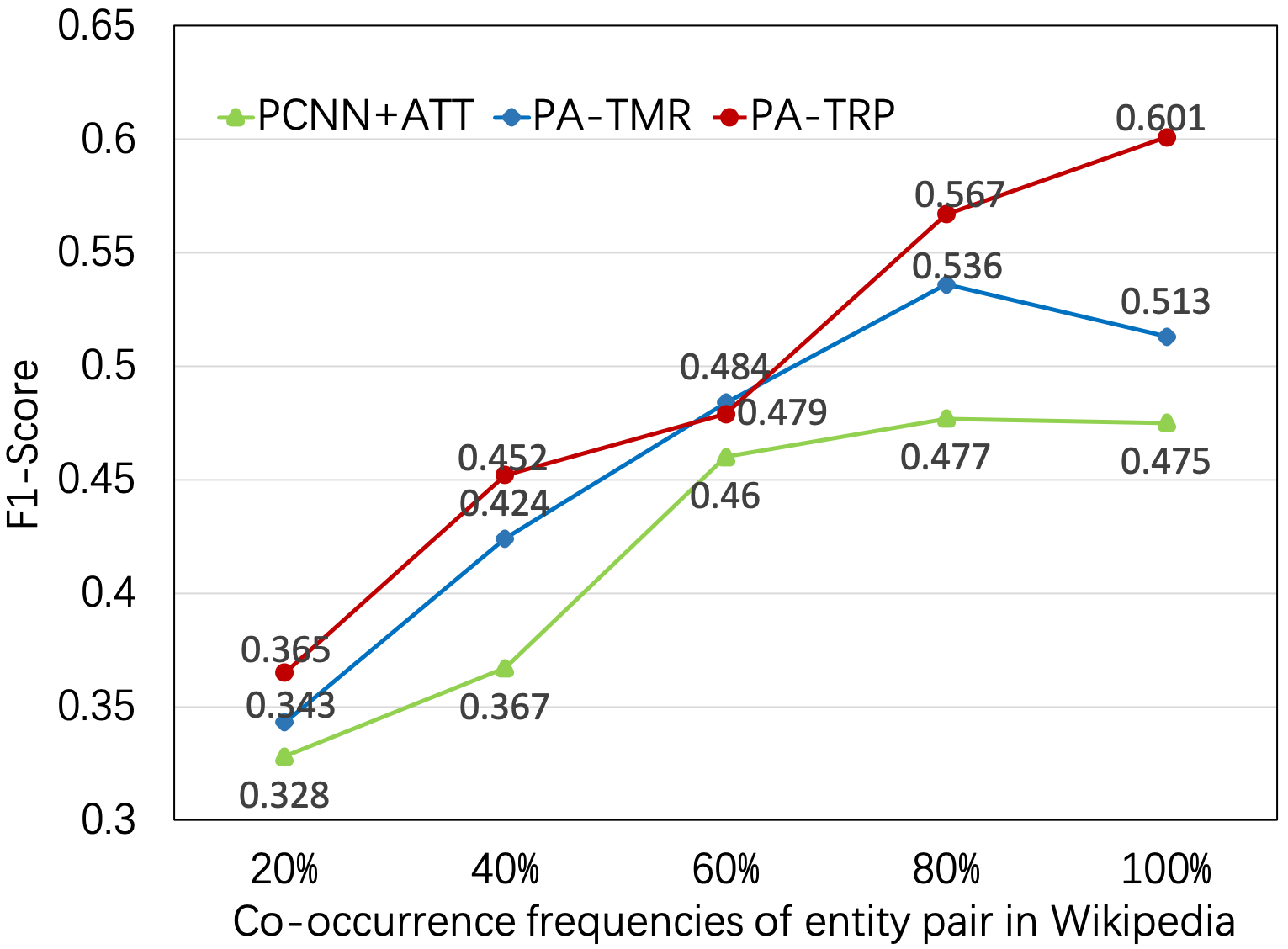}}
  \subfloat[GDS dataset\label{fig:f1_dif_gids}]{
    \includegraphics[width=0.4\linewidth]{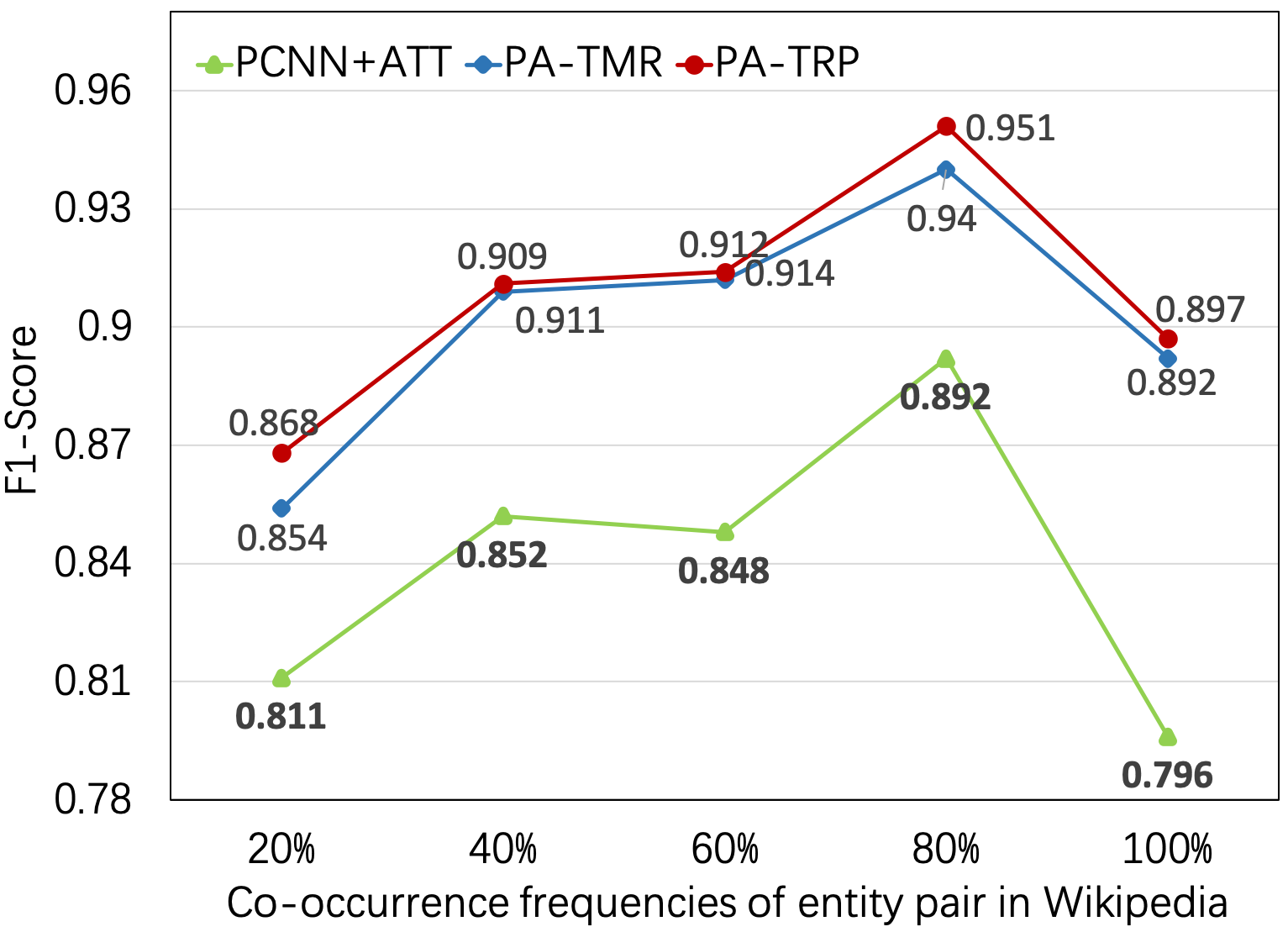}}
  \vspace{-0.1cm}
  \caption{F1-scores with respect to different co-occurrences of entity pair in Wikipedia.}
  \vspace{-0.2cm}
    \label{fig:co_occur}
\end{figure*}

\begin{itemize} [leftmargin=*]
    \item Focusing on NYT, both of our models PA-TMR and PA-TRP outperform all of the other baseline models, mainly because the implicit mutual relations enhance the representations of entity pairs by introducing external text corpora. This helps our model generalize to infrequent or even unseen cases. Compared to PA-TMR, PA-TRP stably performs better, because it further learns the relation prototypes from the referring entity pairs and improves the long-tail relation extraction by transferring knowledge from the proximate relations with sufficient training data.
    \item For GDS, the performance of all the models is fluctuating and unstable. Specifically, when the recall is lower than $0.3$, BGWA and PCNN+HATT achieve the best precisions. The possible reason is that their hierarchical attention mechanism can more efficiently filter out the noisy sentences from the bag, and thus their precisions become higher. When the recall is larger than $0.3$, PA-TMR and PA-TRP outperform other baseline models, mainly because the unlabeled textual information largely benefits recall, while introducing additional noise may decrease precision. PA-TRP is more robust to precision drops when the recall becomes higher (i.e., larger than $0.8$) by capturing the commonality of each relation in their prototypes.
\end{itemize}

Table~\ref{table:result} further demonstrates the above observations with detailed scores. The P@100 and P@200 on the relatively easy dataset GDS almost reach $1.0$, so we pay more attention to the performance on the NYT dataset than that on GDS in the following analysis and ablation studies. We can see:

\begin{itemize} [leftmargin=*]
  \item Our model PA-TRP achieves the best F1-score with $4.6$\% and $3.6$\% performance gains over the second-best models on NYT and GDS, respectively. Particularly, the improvements come from promising recall and competitive precision. Notably, it is demonstrated effective in improving the precision via some advanced mechanisms, such as by filtering out noisy sentences, or by selecting the informative data in the bag. We can see the gradually increasing precisions of the hierarchical attention mechanism in BGWA, the reinforcement learning in CNN-RL, the adversarial training in DSGAN, and the prior relation hierarchy in PCNN-HATT. Moreover, our proposed relation prototypes can be integrated into different RE models for better performance. We leave it in the future, and this paper focuses on the general approach of learning relation prototypes for long-tail relation extraction.
  \item It is also worth pointing out that the precision of PA-TMR consistently outperforms PA-TRP, mainly because PA-TRP does not take the implicit mutual relation between a pair of entities as direct input to the classification layer, which clearly can provide more discriminative signals between proximate relations and benefit the precise classification. 
\end{itemize}

\subsection{Long-tail Relation Extraction (\textbf{RQ2})}
\label{sec:longtail_exp}

In this section, we further study the capability of long-tail relation extraction of our model to investigate the effect of relation prototypes. For fair comparisons, the experiments are conducted on the old version of NYT, because the state-of-the-art baseline models~\cite{zhang2019long,gou2019improving} only report the performance on this dataset, and we cannot find their codes for reimplementation. The old NYT has about another 50,000 training sentences, and the entity pairs in these sentences have overlap with the test dataset. Following baselines, we evaluate the performance of long-tail relations with training instances fewer than $100/200$, and utilize macro top $K$ hit ratio accuracy (Hits@K) as the measurement.

\begin{table}[htbp]
    \centering
    \caption{Overall performance for long-tail relation extraction on NYT. }
    \begin{tabular}{|c|ccc|ccc|}
    \hline
    \#\textbf{Instances} & \multicolumn{3}{c|}{\textless{}\textbf{100}} & \multicolumn{3}{c|}{\textless{}\textbf{200}} \\
    \textbf{Hits@K} & \textbf{10} & \textbf{15} & \textbf{20} & \textbf{10} & \textbf{15} & \textbf{20} \\ \hline
    PCNN+ATT & \multicolumn{1}{c|}{\textless{}5.0} & \multicolumn{1}{c|}{7.4} & 40.7 & \multicolumn{1}{c|}{17.2} & \multicolumn{1}{c|}{24.2} & 51.5 \\ \hline
    PCNN+HATT & \multicolumn{1}{c|}{29.6} & \multicolumn{1}{c|}{51.9} & 61.1 & \multicolumn{1}{c|}{41.4} & \multicolumn{1}{c|}{60.6} & 68.2 \\ \hline
    PCNN+KATT & \multicolumn{1}{c|}{35.3} & \multicolumn{1}{c|}{62.4} & 65.1 & \multicolumn{1}{c|}{43.2} & \multicolumn{1}{c|}{61.3} & 69.2 \\ \hline
    DPEN & \multicolumn{1}{c|}{57.6} & \multicolumn{1}{c|}{62.1} & 66.7 & \multicolumn{1}{c|}{64.1} & \multicolumn{1}{c|}{68.0} & 71.8 \\ \hline\hline

    PA-RP & \multicolumn{1}{c|}{62.0} & \multicolumn{1}{c|}{70.3} & 70.3 & \multicolumn{1}{c|}{65.1} & \multicolumn{1}{c|}{72.3} & 72.3 \\ \hline
    
    PA-TRP & \multicolumn{1}{c|}{\textbf{63.9}} & \multicolumn{1}{c|}{\textbf{70.3}} & \textbf{72.2} & \multicolumn{1}{c|}{\textbf{66.7}} & \multicolumn{1}{c|}{\textbf{72.3}} & \textbf{73.8} \\ \hline
    \end{tabular}
    \label{table:hitsk}

\end{table}

Table~\ref{table:hitsk} shows the overall performance. We can see: (1) the long-tail issue of relation extraction is severe, as the widely used RE model PCNN+ATT has less 5\% Hits@10 accuracy for the relations with less than 100 training instances. (2) Prior relation hierarchy (i.e., PCNN+HATT) can effectively boost the long-tail relation extraction, which benefits from the shared training instances of relations under the same branch, and the incorporation of external KG structures (PCNN+KATT) and entity type information (DPEN) further improve the performance. Especially, the significant improvements on Hits@10 of DPEN demonstrate the importance of entity types. (3) We thus report the performance of both our model and the variant without using type information PA-RP. Both of them achieve satisfactory scores, which demonstrate the effectiveness of our model on addressing the long-tail issue with a general framework of relation prototypes. We will conduct further ablation study by replacing our basic model with other RE models, to verify its generalization ability in Section~\ref{subsection:q3}.

\subsection{The Effect of Implicit Mutual Relations (\textbf{RQ2})}
In this section, we will further investigate the effectiveness of the implicit mutual relations for dealing with the long-tail issue of entity pairs. We consider the settings of entity pair frequency from two aspects.
First, we count them on the introduced unlabeled text corpora and evaluate the performance of our method. Second, we count the co-occurrence frequencies of entity pairs on training data only, aiming to reflect the effect of implicit mutual relations when there is no insufficient training data.

\begin{figure*}[htb]
  \centering
  \subfloat[NYT dataset\label{fig:infreq_nyt}]{
    \includegraphics[width=0.4\linewidth]{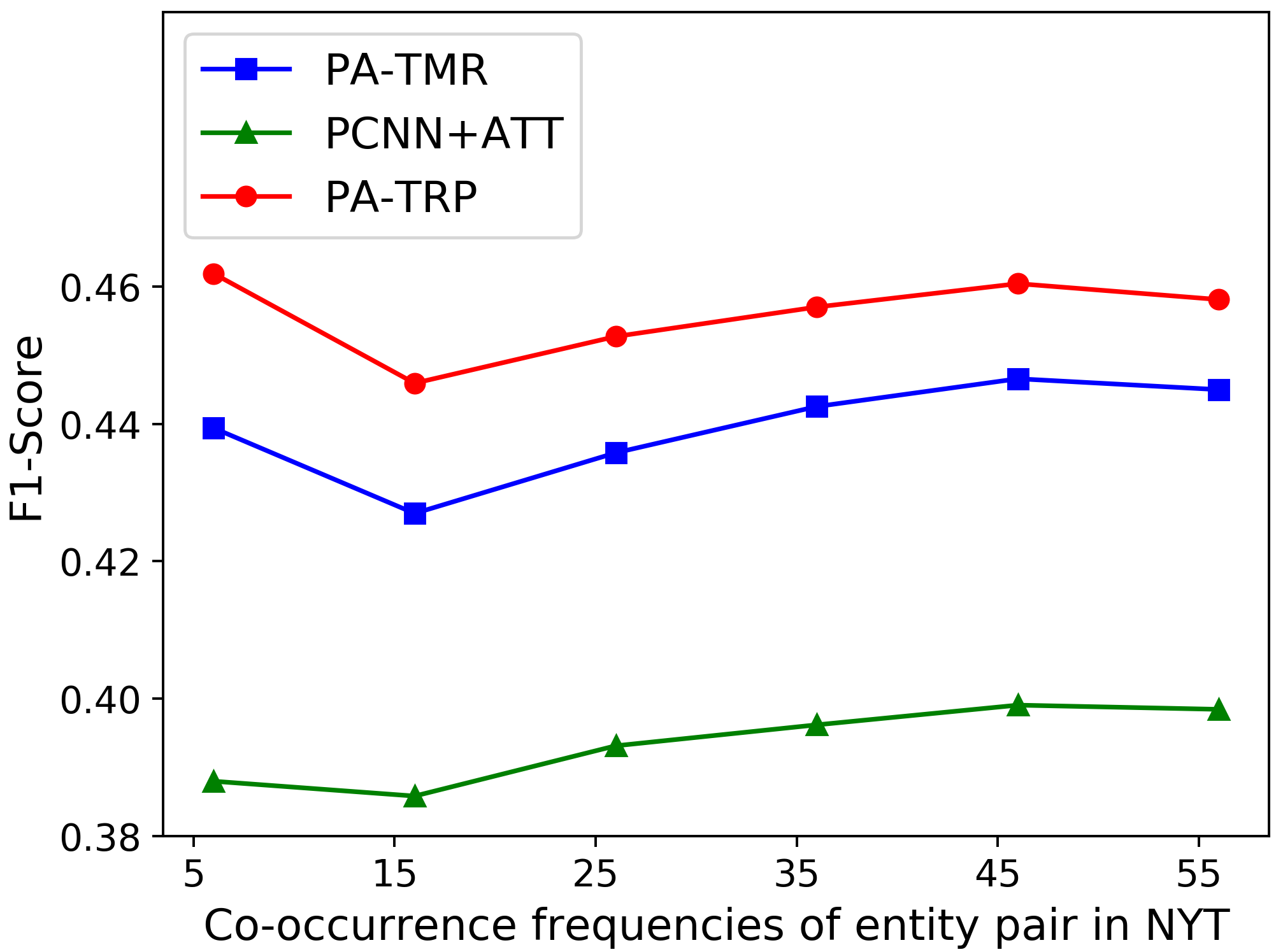}}
  \subfloat[GDS dataset\label{fig:infreq_gds}]{
    \includegraphics[width=0.4\linewidth]{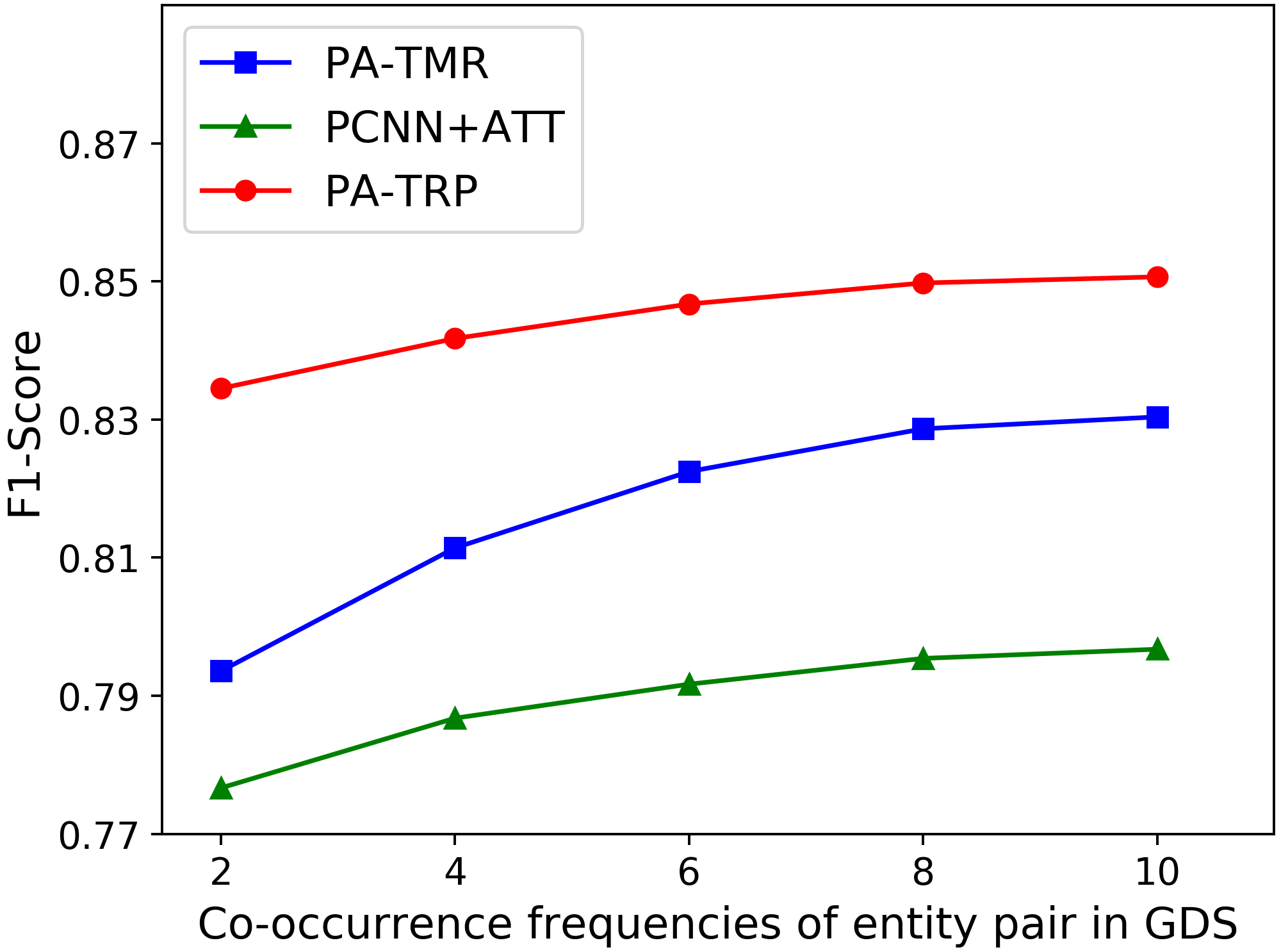}}
  \vspace{-0.1cm}
  \caption{F1-scores with respect to the entity pairs with different co-occurrences in original datasets.}
  \vspace{-0.2cm}
    \label{fig:infreq}
\end{figure*}

\subsubsection{Improvement from implicit mutual relations}

As illustrated in Figure~\ref{fig:co_occur}, we sort entity pairs in ascending order according to their co-occurrences in the unlabeled corpora (Wikipedia). We then evaluate the performance for entity pairs with different co-occurrences, where the x-axis denotes the quantile of co-occurrence times of entity pairs in Wikipedia, and the y-axis denotes the corresponding F1-score. We have the following key observations:
\begin{itemize} [leftmargin=*]
    \item As the co-occurrences of entity pairs increase, F1-score demonstrates an upwards synchronous trend. It reveals that no matter frequent or infrequent entity pairs in the corpora help improve the performance of our PA-TMR and PA-TRP models. This points to the positive effect of all implicit mutual relations collected from the unlabeled corpora. Meanwhile, the implicit mutual relations, which capture the semantic information of both the target entity pair and the entity pairs with similar semantic, contribute to predicting relations for the target entity pair.
    
    \item The improvement on the small dataset GDS is much larger than that on NYT dataset. This is due to the fact that: (1) we insufficiently train the original RE model in the smaller dataset; (2) noisy data in a smaller training dataset exacerbates the inadequate issue of the training process by utilizing the attention mechanism. The better improvement illustrates that implicit mutual relations can alleviate the negative impact of insufficient training corpora.
\end{itemize}

\subsubsection{The impacts of inadequate training sentences}

As illustrated in Figure~\ref{fig:infreq}, we evaluate the impact of inadequate training sentences, where the x-axis denotes the number of training sentences in the distant supervision training corpora, and the y-axis denotes the F1-score of relation extraction. We have the following key observations:
\begin{itemize} [leftmargin=*]
    \item The performance of the original PCNN + ATT increases as an entity pair has more training sentences in the distant supervision training corpora. It reveals that inadequate training sentences indeed have a negative impact on RE.

    \item Our PA-TMR and PA-TRP methods outperform the PCNN+ATT for extracting relations for the entity pairs with inadequate training sentences significantly. This is due to the fact that implicit mutual relations capture extra semantics from the unlabeled text corpora and contribute to predicting their relations.
    
    \item The GDS is much sparser than NYT --- most of the entity pairs have less than ten training sentences. This leads to insufficient training on all entity pairs. The effects of implicit mutual relations become more prominent, along with the increasing frequency. Moreover, the improvements of implicit mutual relations almost stay the same concerning varying training sentences of entity pairs.
\end{itemize}

\subsection{Generalization of Our Method (\textbf{RQ3})}
\label{subsection:q3}

\begin{figure}[htb]
  \centering
  \subfloat[NYT\label{fig:improve-nyt}]{
    \includegraphics[width=0.49\linewidth]{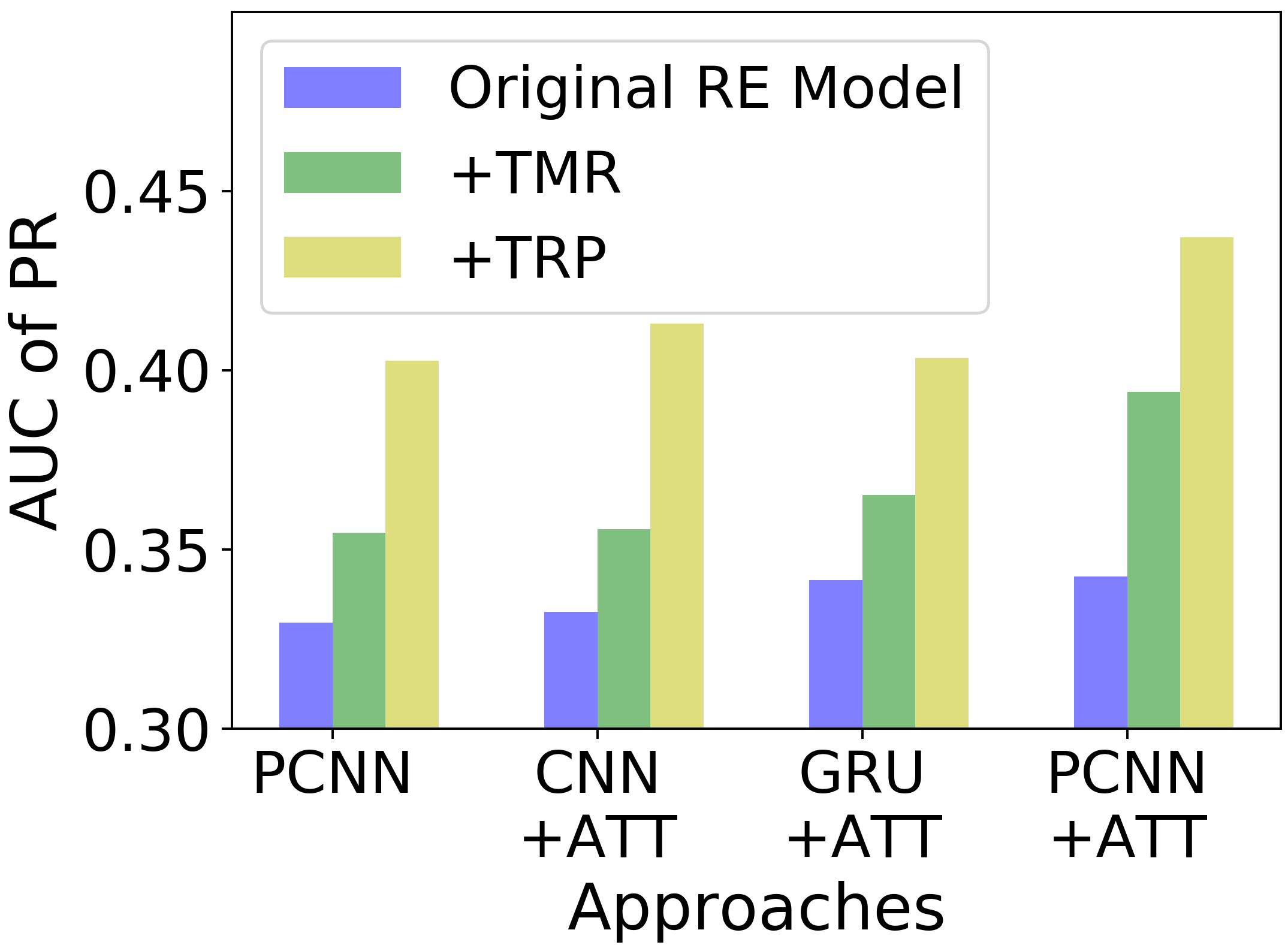}}
  \subfloat[GDS\label{fig:improve-gds}]{
    \includegraphics[width=0.49\linewidth]{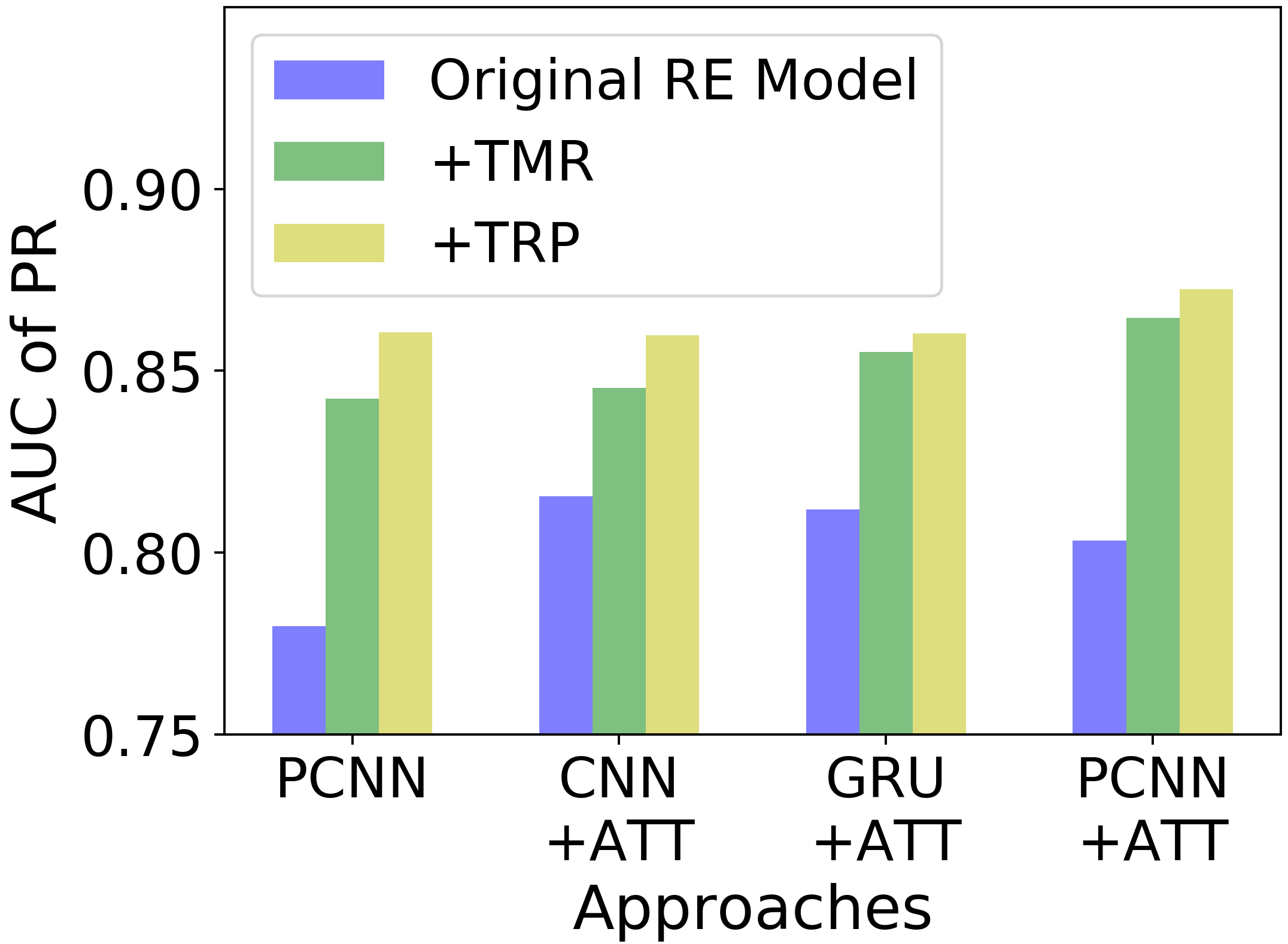}}
  \vspace{-0.1cm}
  \caption{Generalization ability of relation prototypes to various RE models. }
  \vspace{-0.2cm}
    \label{fig:improve}
\end{figure}

To illustrate the flexibility of our relation prototypes and implicit mutual relations, we incorporate them into different neural relation extraction approaches, including GRU based model with sentence-level attention (GRU+ATT), CNN + ATT~\cite{senatt}, PCNN~\cite{pcnn}, and PCNN + ATT~\cite{senatt}. As elaborated in Figure~\ref{fig:improve}, we have the following key observations:

\begin{itemize} [leftmargin=*]
    \item The performance of original RE models varies a lot on the two datasets. Among them, the attention mechanism plays a critical role, no matter in which datasets and in which types of neural networks. In terms of the neural encoder, PCNN outperforms than GRU and CNN on NYT, while, the vanilla CNN performs better on GDS.
    \item The incorporation of both TMR and TRP achieves significant improvements across all of these models. Especially, we stably have the most performance gains by using PCNN+ATT as our basic neural encoder on two datasets. This demonstrates the potential of PCNN that models entity mentions separately.
    \item Compared to TMR, the improvements of TRP is larger on NYT than that on GDS. This is mainly because, on GDS, the transferring learning among relations is less important than the enhancements of entity pair representations with implicit mutual relations. There are only five relations to classify (Table~\ref{table:dataset}), and each relation has sufficient training instances. However, GDS still suffers the long-tail issue with respect to entity pairs. This demonstrates the effectiveness of our proposed method to deal with the long-tail issues from both two aspects: the relations with insufficient training instances and infrequent or even unseen entity pairs.
\end{itemize}

\subsection{Impacts of Each Components}

\begin{figure}
   \centering
  \subfloat[NYT\label{fig:nyt-each}]{
    \includegraphics[width=0.49\linewidth]{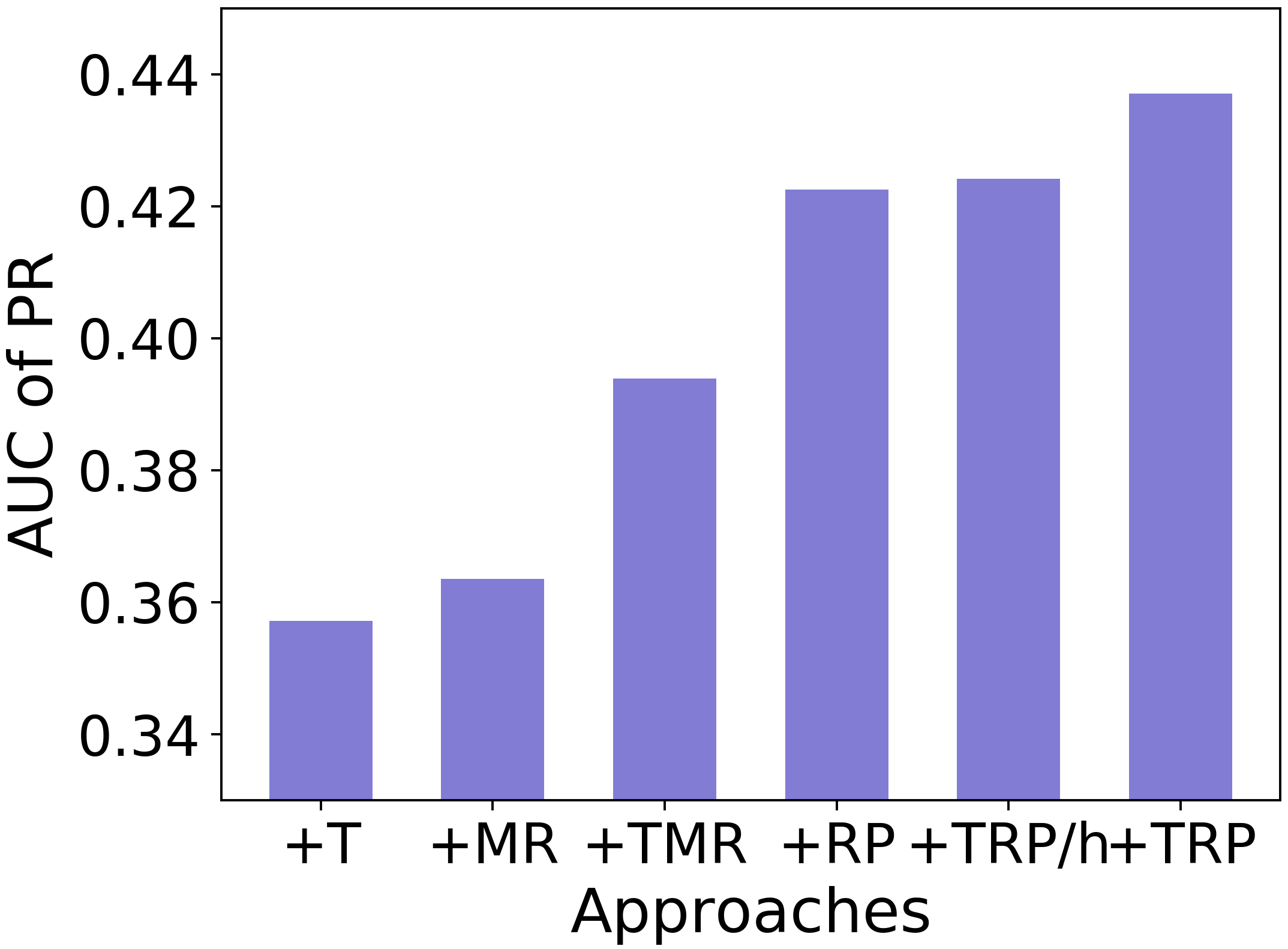}}
  \subfloat[GDS\label{fig:gds-each}]{
    \includegraphics[width=0.49\linewidth]{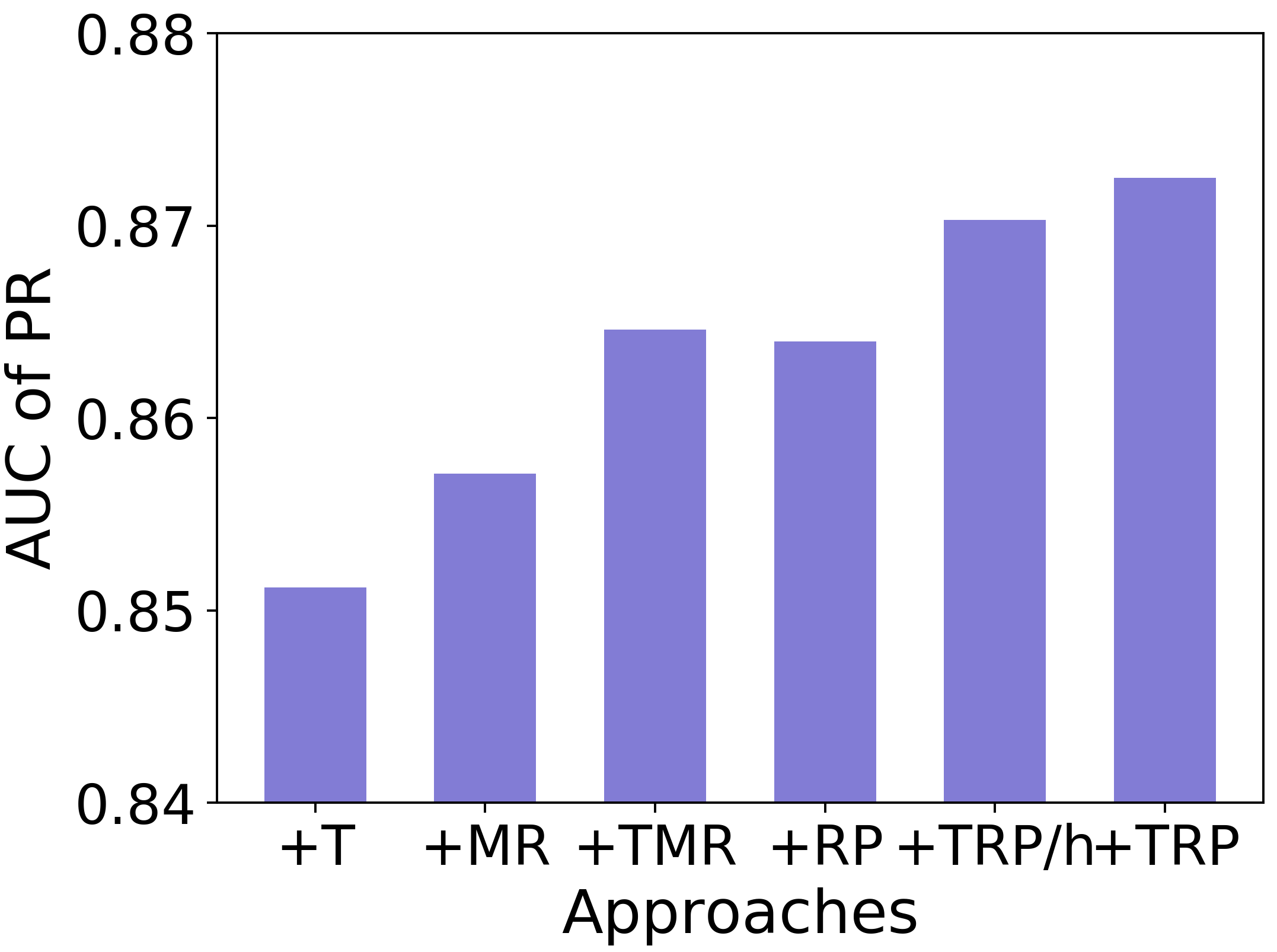}}
  \caption{Impacts of each component by removing them from PA-TRP one by one and presenting their ACU scores.}
    \label{fig:eachcom}
\end{figure}

To verify the effectiveness of each main component in our proposed model, we conduct an ablation study by removing each of them and present their performance in Figure~\ref{fig:eachcom}. We can see: (1). the utilization of the type (i.e., +T) or implicit mutual relations between pair of entities (i.e., +MR) can only bring limited performance gains, while a noticeable increase of the combination of them (i.e., +TMR) demonstrates that they are highly complementary. (2). The learned relation prototypes (i.e., +RP) boost the performance compared to using implicit mutual relations only (i.e., +MR) on both datasets. They actually use the same information, since relation prototypes are derived from the centroid of the referred entity pairs. (3). By incorporating the prior relation hierarchy (i.e., +TRP/h v.s. +TRP), our proposed model further improves the performance on NYT, and the improvements on GDS is slight because of the limited prior information among five relations only.

\subsection{Case Study}
In this section, we conduct qualitative analysis by presenting actual cases to give an intuitive impression of our method. Still, we present cases from two perspectives of relation prototypes and implicit mutual relations.

\begin{table}[htbp]
  \centering
\caption{Top 5 proximate relations based on prototypes given entity pairs \textit{(Namibia, Windhoek)} and \textit{(General Motors, Flint)}.}
\label{table:rp_case}
    \scalebox{0.8}{
        \begin{tabular}{|c|c|c|}
        \hline
        \textbf{Entity Pair} & \textbf{Top 5 proximate relations} & \#\textbf{Training} \\ \hline
        \multirow{5}{*}{\begin{tabular}[c]{@{}c@{}}(Namibia, \\ Windhoek)\end{tabular}} & \textbf{/location/country/capital} & 170\\ \cline{2-3} 
         & /location/country/administrative\_divisions & 389 \\ \cline{2-3} 
         & /location/location/contains & 8321 \\ \cline{2-3} 
         & /location/fr\_region/capital & 1 \\ \cline{2-3} 
         & /location/cn\_province/capital & 2 \\ \hline \hline
        \multirow{5}{*}{\begin{tabular}[c]{@{}c@{}}(General Motors, \\ Flint)\end{tabular}} & /bussiness/company/locations & 3 \\ \cline{2-3} 
         & \textbf{/business/company/place\_founded} & 179 \\ \cline{2-3} 
         & /business/company/major\_shareholders & 29 \\
         \cline{2-3} 
         &/time/event/locations & 2 \\
         \cline{2-3} 
         & /film/film\_festival/location & 4 \\ \hline
         
        \end{tabular}
    }
\end{table}

\subsubsection{Proximity among relation prototypes}
Relation prototypes aim to capture the meaning of relations and their proximity relationships. Given any entity pair, we can compute its similarity with these prototypes to provide possible relation candidates and transfer knowledge among proximate relations. We adopt cosine similarity as measurement. As demonstrated in Table~\ref{table:rp_case}, we report top 5 proximate relations with the highest scores to entity pairs \textit{(Namibia, Windhoek)} and \textit{(General Motors, Flint)}. Correct relations are denoted as bold fonts.

We can see that the relation prototypes already can provide a strong baseline for ranking correct relations pretty high (i.e., $1$st and $2$nd). Besides, all of the neighbors are proximate even they do not belong to the same branch, which provides complementary signals to the prior relation hierarchy and benefits the knowledge transfer. This demonstrates the effectiveness of our relation prototype learning based on implicit mutual relations. Next, we also wonder how the implicit mutual relations of entity pairs are well coherent with each other if they share similar relations, in order to ensure the high quality of relation prototypes and alleviate the negative impacts of insufficient training of infrequent or even unseen entity pairs.

\subsubsection{Similarity among implicit mutual relations}

\begin{table}[h]
\centering
\caption{Top 10 similar implicit mutual relations given entity pair \textit{(Stanford University, California)}}
\label{table:cosine}
\begin{tabular}{|c|c|c|}
\hline
\textbf{Entity pairs} & \textbf{Cosine similarity} & \textbf{Relation} \\ \hline
(University of Chicago, Chicago) & 0.788 & \textit{locatedIn} \\ \hline
\begin{tabular}[c]{@{}c@{}}(University of Southern California, \\ Los Angeles)\end{tabular} & 0.758 & \textit{locatedIn} \\ \hline
(Mikheil Saakashvili, Tbilisi) & 0.697 & \textit{bornIn} \\ \hline
\begin{tabular}[c]{@{}c@{}}(Columbia University, \\ New York City)\end{tabular} & 0.687 & \textit{locatedIn}\\ \hline
(University of London, London) & 0.681 & \textit{locatedIn} \\ \hline
(Abilene,Texas) & 0.681 & \textit{locatedIn}\\ \hline
\begin{tabular}[c]{@{}c@{}}(University of Pennsylvania, \\ Philadelphia)\end{tabular} & 0.668 & \textit{locatedIn}\\ \hline
(Chapel Hill, North Carolina) & 0.667 &\textit{locatedIn}\\ \hline
(Stanford, Florida) & 0.665 &\textit{locatedIn}\\ \hline
(Central Florida, Florida) & 0.658&\textit{locatedIn} \\ \hline
\end{tabular}
\end{table}

Remember that the implicit mutual relation is represented as the difference between the pair of entity embeddings, which are learned from the entity co-occurrence graph. Thus, given any entity pair, e.g., \textit{(Stanford University, California)}, we can compute the nearest neighbors according to their implicit mutual relations. We adopt cosine similarity as the measurement in the embedding space. As demonstrated in Table~\ref{table:cosine}, we report the top 10 neighbors with the highest cosine scores. If cosine similarity of the embedding offsets is high, the corresponding entity pair is similar, and they tend to have similar relations. We can observe that only one entity pair has different relations to (Stanford University, California), and most entity pairs share the same ``\textit{locatedIn}'' relation. It indicates that the defined entity co-occurrence graph is reasonable to capture the implicit mutual relations after vertex embedding. Furthermore, the implicit mutual relations of entity pairs are close to each other if they have the same relation, which benefits the prototype learning.

\section{RELATED WORK}~\label{sec:rw}

Recently, neural network methods for relation extraction have made great progress. 
Zeng et al.~\cite{cnn} propose a CNN-based model which can capture features at both lexical and sentence levels. PCNN~\cite{pcnn} introduces piecewise max-pooling in CNN to separate the textual features from head entity, tail entity, and relations. To further improve the performances, a host of work focuses on the following aspects: advanced Neural Encoder, side Information, and robustness to Poor Annotations.
Next, we will describe the main progress in each aspect.

\subsection{Advanced Neural Encoder}
The neural encoder that learns text representations plays a critical role in relation extraction. A better neural encoder with strong feature abstraction ability will result in superior performance~\cite{walked,miwa2016end}. Therefore, many works focus on improving the neural encoder to get more prominent relation extraction models. Santos et al.~\cite{crcnn} propose a ranking-based convolutional neural network (CR-CNN), and Nguyen et al.~\cite{nguyen} utilize multiple window sizes for CNN filters to obtain features from various granularities. 
Miwa et al.~\cite{miwa2016end} stack bidirectional tree-structured LSTM-RNNs on bidirectional sequential LSTM-RNNs to encode both word sequence and dependency tree.
To encode multiple entity pairs in a sentence simultaneously, Christopoulou et al.~\cite{walked} regard entities as nodes in a fully-connected graph and encode them with a walk-based model.

Recently, pre-trained language models (LMs), such as BERT~\cite{bert}, can provide a powerful neural encoder and achieve great success in many downstream tasks. However, although they perform well on sentence-level RE~\cite{soares2019matching}, few studies apply them to bag-level RE. This is because LMs have a promising finetuning performance with only a few training data, while they suffer from inefficiency. Bag-level RE datasets can utilize distant supervision to collect adequate annotations for training, but inevitably introduce much noise. Moreira et al.~\cite{moreira2020distantly} presents unsatisfactory Precision-Recall curves of BERT-based RE. Yu et al.~\cite{yu2020relation} designs a time-decay selective attention mechanism to deal with the noisy issue, and achieves improvements. That is, how to efficiently model the long-tail bags of sentences and mitigate the negative impacts of noise become key challenges of bag-level RE. More advanced neural encoder is not our main focus. Instead, we aim to provide a general way to transfer knowledge between proximate relations, which can be applied to various neural architectures.

\subsection{Side Information}

There is some external knowledge that benefits the relation classification, such as relation alias information~\cite{reside2018}, part-of-speech tags~\cite{postag} and semantic information~\cite{D18-1201}. Therefore, some works tend to introduce this useful information as additional supervision signals for superior performance.

For relation information, Vashishth et al.~\cite{reside2018} utilize the relation alias information (e.g. \textit{founded} and \textit{co-founded} are aliases for the relation \textit{founderOfCompany}) to enhance the relation representation. Zeng et al.~\cite{relationpath} construct the relation path to facilitate longer dependency between entities, which may be not in the same sentence.
For entity information, Ji et al~\cite{ji2017distant} introduce entity descriptions to enhance entity representations, while Liu et al.~\cite{entitytype} utilize entity type information, which is directly related to specific relations. For example, the relation \textit{capital} must be between two location entities, rather than between two persons.
Another interesting work directly employs the KG as the side information~\cite{distiawan2019neural}. They utilize an end-to-end neural network to complete the missing relations in KG and to reduce error propagation between relation extraction and the upstream task named entity disambiguation.
Besides, additional information from related tasks are demonstrated to be effective, such as NER~\cite{li2019adversarial,li2020metaner,li2020survey,li2020neural} and correlation analytics~\cite{xie2013local}.

The quality and quantity of the above-mentioned information highly depend on their sources, which can not always be guaranteed. Therefore, we focus on the unlabeled text that extensively exists on the web, mine implicit mutual relations between entities, and utilize them to represent relation prototypes for transfer learning. Furthermore, our proposed method can also employ other additional information (i.e., types in experiments) for better performance.

\subsection{Robustness to Poor Annotations}
Neural network models usually require a large amount of training data, which is expensive to obtain or only available in specific domains. This leads to the problem of lacking annotations, which become more serious for large-scale datasets. To address the issue, the distant supervision is proposed~\cite{ds} to annotate relation in texts automatically. The assumption is that if an entity pair $(e_h, e_t)$ has a relation $r$, any sentence that contains $e_h$ and $e_t$ might express that relation. So labeled data can be obtained by aligning sentences to KGs. However, the distant supervision will inevitably introduce noise and bring long-tail problems (as discussed in Section~\ref{sec:intro}). Therefore, many works attempt to address how to alleviate the performance loss caused by noisy data~\cite{senatt,hoffmann} and by the relations without sufficient training corpora~\cite{han2018hierarchical,zhang2019long,gou2019improving}. We roughly classify them into two groups in terms of the noise and long-tail problem.

\subsubsection{Denoise in distant supervision}

To mitigate the noise caused by distant supervision, some works~\cite{riedel}\cite{hoffmann} utilize multi-instance learning, which allows different sentences to have at most one shared label. The multi-instance learning combines all relevant instances to determine the relation of the targeted entity pair, which thereby alleviates the impact of incorrectly labeled instances. Surdeanu et al.~\cite{surdeanu} get rid of the restrict that different sentences can only share one label by utilizing a graphical model. It can jointly model multiple instances and relations. 

With the development of neural network, attention mechanism is proposed to help neural models focus on important training data. In the field of relation extraction, attention mechanism is widely used to mitigate the effects of noisy data~\cite{P1183-1194}. Existing attention approaches can be categorized into two groups: sentence-level attention and word-level attention. Sentence-level attention~\cite{senatt} aims at selecting the sentences w.r.t., the relational strength between the target entity pair. Similarly, word-level attention~\cite{wordatt} focuses on high-quality words to measure the target relation.
Furthermore, Wang et al.~\cite{wang2016relation} adopt hierarchical attention, which combines these two attention mechanisms. 

Alternatively, reinforcement learning can also alleviate the effects of noisy data~\cite{reinforce}~\cite{Qin2018RobustDS}. The reinforcement learning methods mainly consist of two modules: an instance selector module aims to select the high-quality instances, and the other module of relation classifier makes the prediction and provides a reward to the instance selector. The noisy data will be eliminated by the instance selector, leading to a performance gain.

Adversarial training~\cite{goodfellow2014explaining} is also a viable solution to address the noise problem. Wu et al.~\cite{adversarial} generate adversarial samples by adding noise of small perturbations to the original data, then encourage the neural network to correctly classify both unmodified examples and perturbed ones for regularization. The resulting model becomes more robust and generalizable. Furthermore, Qin et al.~\cite{qin2018dsgan} utilize the Generative Adversarial Networks(GANs)~\cite{gans} to filter distant supervision training dataset and redistribute the false positive instances into the negative training set.

\subsubsection{Long-tail relation extraction}
Although many works aim to alleviate the negative impacts of noise, only a few works focus on improving long-tail relation extraction. Some recent works highlight the importance of RE in few shot settings~\cite{han2018fewrel,ye2019multi} and achieve great success. Differently, long-tail RE naturally includes relation types with different numbers of training sentences, thus focuses on incorporating prior relation hierarchy for knowledge transfer~\cite{han2018hierarchical,zhang2019long}. The relations under the same branch are regarded as closely correlated, and can take advantage of each other's training data. DPEN~\cite{gou2020dynamic} incorporates entity type information to learn relation-specific classifier dynamically.

In this paper, we aim to fill this blank of improving long relation extraction. Especially, we try to minimize the reliance on additional information, such as prior hierarchy and entity types. Instead, we only utilize unlabeled texts, which are easy to obtain. We aim to capture the commonality among relations for knowledge transfer and the differences between entity pairs for discrimination.

\section{CONCLUSION and FUTURE WORK}~\label{sec:conclusion}

In conclusion, we have proposed a general approach to learn relation prototypes from unlabeled texts. The prototype learning method can be applied in current models for better relation extraction by transferring knowledge from relations with sufficient training data to long-tail relations. We have conducted extensive experiments to verify the effectiveness of the proposed method on two publicly available datasets and compared them with eight state-of-the-art baselines. The results present significant improvements, especially in long-tail settings.
Further ablation study and case study also demonstrate the effectiveness of our proposed method and the generalization ability to current RE models from both quantitative and qualitative perspectives.
In the future, we are interested in enhancing entity embeddings with KG including structure and attribute information.

%investigating more advanced entity embedding models, such as Graph Attention Networks (GATs)~\cite{gat}, to improve the implicit mutual relation representation as well as relation prototypes. Also, other side information (e.g., Knowledge Graph including structural and numerical information) can be incorporated to enrich the entity co-occurrence graph for better modeling.

\section*{Acknowledgments}
\label{sec:acknowledge}
This work has been supported by the National Key Research and Development Program of China under grant 2016YFB1000905
%(Ming Gao)
, and the National Natural Science Foundation of China under Grant No. U1811264
%(cheqing jin)
, 61877018
%(ming gao)
, 61672234
%(qiwendong)
, 61672384
%(Wei Wang)
, and the Shanghai Agriculture Applied Technology Development Program, China (Grant No.T20170303).
% or


% \appendices

% \section{}

% \ifCLASSOPTIONcompsoc
  % \section*{Acknowledgments}
% \else
  % \section*{Acknowledgment}
% \fi


\ifCLASSOPTIONcaptionsoff
  \newpage
\fi



\bibliographystyle{IEEEtran}
\bibliography{reference}


% \end{thebibliography}

\begin{IEEEbiography}[{\includegraphics[width=1in,height=1.24in,clip,keepaspectratio]{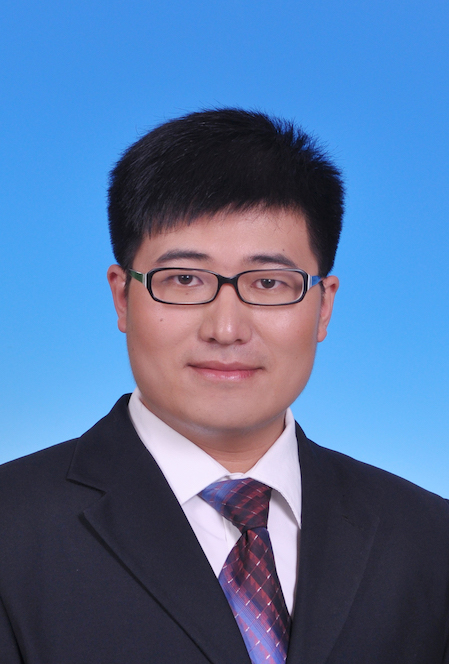}}]{Yixin Cao} is a research assistant professor with Nanyang Technology University. Before that, He was a research fellow with National University of Singapore. He received his Ph.D. from Tsinghua University. His research interests span NLP and KG. Various parts of his work have been published in top conferences, such as ACL, EMNLP, AAAI, and WWW. Moreover, he has served as PC/SPC for conferences including ACL, EMNLP, AAAI, and IJCAI, and invited reviewers for journals including TKDE, TKDD, and IEEE Access.
\end{IEEEbiography}

\begin{IEEEbiography}[{\includegraphics[width=1in,height=1.24in,clip,keepaspectratio]{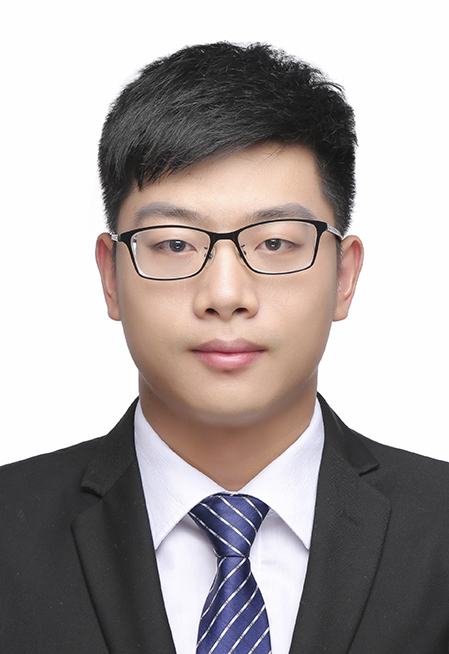}}]{Jun Kuang} is currently a master student on School of Data Science and Engineering at East China Normal University, China. His research interests include knowledge engineering and nature language processing. His work appears in the International Conference on Data Engineering.

\end{IEEEbiography}

\begin{IEEEbiography}[{\includegraphics[width=1in,height=1.24in,clip,keepaspectratio]{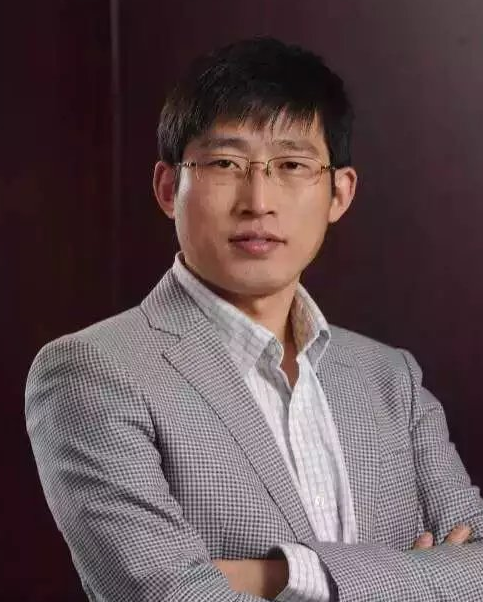}}]{Ming Gao}
is currently a full professor on School of Data Science and Engineering at East China Normal University, China. He received his doctorate from the School of Computer Science, Fudan University.
%after that he conducted three-year postdoctoral research at Singapore Management University for the project LiveLabs.
%
His research interests include knowledge engineering, user profiling, social network analysis and mining. His works appear in major international journals and conferences, including TKDE, KAIS, DMKD, ICDE, SIGIR, ICDM, etc.
\end{IEEEbiography}

\begin{IEEEbiography}[{\includegraphics[width=1in,height=1.24in,clip,keepaspectratio]{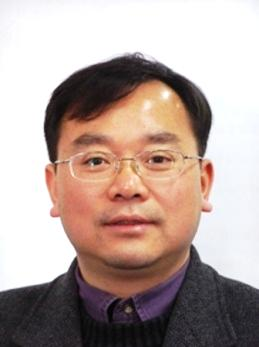}}]{Aoying Zhou}
is a full professor on School of Data Science and Engineering (DaSE) at East China Normal University (ECNU), where he is the founding dean of DaSE and the vice president of ECNU. He is the winner of the National Science Fund for Distinguished Young Scholars supported by NSFC and the professorship appointment under Changjiang Scholars Program of Ministry of Education. His research interests include Web data management, data intensive computing, in-memory cluster computing and benchmark for big data. His works appear in major international journals and conferences, including TKDE, PVLDB, SIGMOD, SIGIR, KDD, WWW, ICDE and ICDM, etc.
\end{IEEEbiography}

\begin{IEEEbiography}[{\includegraphics[width=1in,height=1.24in,clip,keepaspectratio]{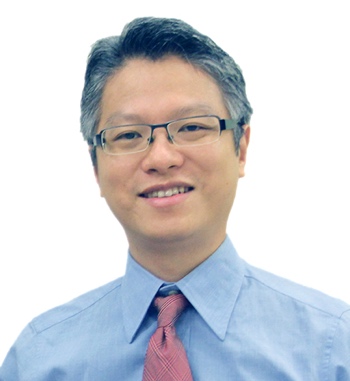}}]{Yonggang Wen}
  is a Professor with the School Of Computer Engineering, Nanyang Technological University, Singapore. His research interests include cloud computing, green data centers, big data analytics, multimedia networks, and mobile computing. His latest work in multiscreen cloud social televisions has been featured by global media and recognized with ASEAN ICT Award 2013 and IEEE Globecom 2013 Best Paper Award. He serves on the editorial boards of IEEE TRANSACTIONS ON MULTIMEDIA, IEEE ACCESS JOURNAL, and Elsevier Ad Hoc Networks.
  \end{IEEEbiography}

\begin{IEEEbiography}[{\includegraphics[width=1in,height=1.24in,clip,keepaspectratio]{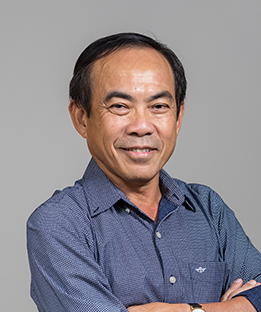}}]{Tat-Seng Chua}
is the KITHCT Chair Professor at the School of Computing, National University of Singapore. He was the Acting and Founding Dean of the School from 1998. Dr Chua's main research interest is in multimedia information retrieval and social media analytics. He is the co-Director of NExT, a joint Center between NUS and Tsinghua University. Dr Chua is the 2015 winner of the prestigious ACM SIGMM award for Outstanding Technical Contributions to Multimedia Computing, Communications and Applications. He is the Chair of steering committee of ACM ICMR and MMM. Dr Chua is also the General Co-Chair of ACM Multimedia'05, ICMR'05, SIGIR'08, and Web Science'15. He serves in the editorial boards of four international journals.
\end{IEEEbiography}

% \end{IEEEbiography}


%\newpage



%\vfill


\end{document}